\documentclass[10pt]{article}
\usepackage[T1]{fontenc}
\usepackage{iftex}
\ifPDFTeX
  \usepackage[utf8]{inputenc}
\fi
\usepackage[margin=1in]{geometry}
\usepackage{amsmath,amssymb,booktabs,array,graphicx,xcolor}
\usepackage{newtxtext,newtxmath}
\usepackage[
sort&compress]{natbib}
\setcitestyle{round}
\usepackage[hidelinks]{hyperref}
\usepackage{microtype}
\usepackage{float}
\usepackage[font=small,labelfont=bf,labelsep=period,skip=4pt]{caption}
\usepackage{placeins}
\usepackage{tikz}
\usepackage{makecell}
\usepackage{multirow}
\usetikzlibrary{arrows.meta,calc,positioning,fit,backgrounds}
\definecolor{metablue}{RGB}{34,102,188}
\newcommand{\pih}{\pi_h}
\newcommand{\pip}{\pi_p}
\newcommand{\piz}{\pi_0}
\newcommand{\pij}{\pi_{hp}}
\DeclareRobustCommand{\method}{BLaDe}

\usepackage{xcolor}

\title{
Bayesian Language Decision: Looking Back and Planning Ahead
}
\author{%
Haicheng Huang\textsuperscript{*}\quad Lang Lin\textsuperscript{*}\quad Xinyang Liu\textsuperscript{*}\quad
Yanbo Wang\quad Tianyu Chen\quad Ruyi An\quad Mingyuan Zhou\textsuperscript{\(\dagger\)}\\[2pt]
\small The University of Texas at Austin\\[1pt]
\footnotesize\textsuperscript{*}Equal contribution.\quad
\textsuperscript{\(\dagger\)}Corresponding author.}
\date{}

\usepackage{titlesec}
\usepackage{fancyhdr}

\definecolor{JanusNavy}{RGB}{0,0,128}
\definecolor{JanusAbstractFill}{rgb}{0.90,0.95,1.00}
\definecolor{JanusCitationBlue}{RGB}{0,112,160}
\definecolor{JanusCrossReference}{RGB}{128,0,0}

\titleformat{\section}
  {\color{JanusNavy}\normalfont\fontsize{13}{15.6}\selectfont\bfseries}
  {\thesection.}{0.55em}{}
\titleformat{\subsection}
  {\color{JanusNavy}\normalfont\fontsize{11}{13.2}\selectfont\bfseries}
  {\thesubsection.}{0.55em}{}
\titleformat{\subsubsection}
  {\color{JanusNavy}\normalfont\normalsize\bfseries}
  {\thesubsubsection.}{0.55em}{}
\titlespacing*{\section}{0pt}{8pt plus 2pt minus 2pt}{4pt}
\titlespacing*{\subsection}{0pt}{6pt plus 2pt minus 2pt}{3pt}
\titlespacing*{\subsubsection}{0pt}{5pt plus 2pt minus 1pt}{3pt}
\titlespacing*{\paragraph}{0pt}{4pt plus 1pt minus 1pt}{0.5em}

\DeclareCaptionLabelSeparator{janusbar}{\enspace|\enspace}
\hypersetup{colorlinks=true,linkcolor=JanusCrossReference,
  citecolor=JanusCitationBlue,urlcolor=JanusCitationBlue}

\makeatletter
\def\janus@displaytitle#1:#2\@nil{%
  {\upshape\bfseries #1}{\upshape\mdseries:#2}}
\let\janus@runningtitle\@title
\fancypagestyle{plain}{%
  \fancyhf{}%
  \fancyfoot[C]{\small\thepage}%
}

\renewcommand{\@maketitle}{%
  \newpage
  \begingroup
  \let\footnote\thanks
  \setlength{\parindent}{0pt}%
  \vspace{2pt}\hrule height 0.6pt\vspace{8pt}%
  \raggedright
  {\color{JanusNavy}\fontsize{21}{23}\selectfont
    \expandafter\janus@displaytitle\@title\@nil\par}%
  \vspace{7pt}%
  \begingroup
    \renewcommand{\small}{\fontsize{8.5}{10.2}\selectfont\mdseries}%
    \renewcommand{\footnotesize}{\fontsize{8}{9.6}\selectfont\mdseries}%
    {\fontsize{9.5}{11.5}\selectfont\bfseries
      \begin{tabular}[t]{@{}l@{}}\@author\end{tabular}\par}%
  \endgroup
  \vspace{7pt}\hrule height 0.6pt\vspace{4pt}%
  \endgroup
}
\makeatother

\newsavebox{\JanusAbstractBox}
\newlength{\JanusAbstractTextWidth}
\newlength{\JanusAbstractPadding}
\renewenvironment{abstract}{%
  \par\addvspace{4pt}%
  \setlength{\JanusAbstractTextWidth}{\dimexpr\linewidth-2\JanusAbstractPadding\relax}%
  \begin{lrbox}{\JanusAbstractBox}%
  \begin{minipage}{\JanusAbstractTextWidth}%
    \small\setlength{\parindent}{0pt}\setlength{\parskip}{0pt}%
    \noindent\textbf{\abstractname}\enspace\ignorespaces
}{%
    \par\end{minipage}%
  \end{lrbox}%
  \noindent\begin{tikzpicture}
    \node[fill=JanusAbstractFill,draw=none,rounded corners=3pt,
      inner sep=\JanusAbstractPadding,outer sep=0pt]
      {\usebox{\JanusAbstractBox}};
  \end{tikzpicture}\par\addvspace{4pt}%
}

\let\janusBibliography\thebibliography
\renewcommand{\thebibliography}[1]{%
  \janusBibliography{#1}%
  \small
  \setlength{\itemsep}{1pt}%
  \setlength{\parskip}{0pt}%
  \setlength{\parsep}{0pt}%
}

\AtBeginDocument{%
  \setlength{\textfloatsep}{6pt plus 2pt minus 2pt}%
  \setlength{\floatsep}{6pt plus 2pt minus 2pt}%
  \setlength{\intextsep}{6pt plus 2pt minus 2pt}%
}

\newcounter{algorithm}
\newfloat{algorithmfloat}{t}{loa}
\newenvironment{metalgorithm}[1]{%
\begin{algorithmfloat}[!t]\noindent\begin{minipage}{\linewidth}%
\refstepcounter{algorithm}\hrule\smallskip%
\textbf{Algorithm \thealgorithm: #1}\par\smallskip\hrule\smallskip\small%
}{\smallskip\hrule\end{minipage}\end{algorithmfloat}}

\begin{document}
\maketitle

\begin{abstract}
System~I decision models select among candidate actions directly, whereas System~II models allocate additional inference-time computation to deliberative planning. Pairing the two lets a fast actor pursue long-horizon plans without invoking the planner at every step. Between planner calls, however, each decision must weigh what recent interactions reveal against where the plan points. We treat this trade-off as Bayesian inference over actions, with the past forming a prior and the planned future serving as evidence. Scoring this evidence directly would require knowing where each action leads. Under a simple factorization, Bayes' rule instead recasts it as how much the planned future raises or lowers each action's probability under the actor itself. We introduce \textbf{B}ayesian \textbf{La}nguage \textbf{De}cision (\textbf{\method{}}), a training-free framework that queries one frozen actor under three contexts (the observation alone, with recent history, and with the planner's target) and combines the answers through a single weight $w$ that sets how far the plan may override the past.
Across 
three System~I models and
four interactive benchmarks, \method{} improves success over the actor alone in every actor--environment pair,
raising average success from 14.76\% to 40.38\% at $w=2$ and cutting environment actions on ALFWorld household tasks by up to 52.5\%, all without updating the actor.
\end{abstract}

\begin{figure}[H]
\centering
\newif\ifJanusMainResultsPDF
\IfFileExists{figures/main_results_bar.pdf}{\JanusMainResultsPDFtrue}{\JanusMainResultsPDFfalse}
\ifJanusMainResultsPDF
\includegraphics[width=\linewidth]{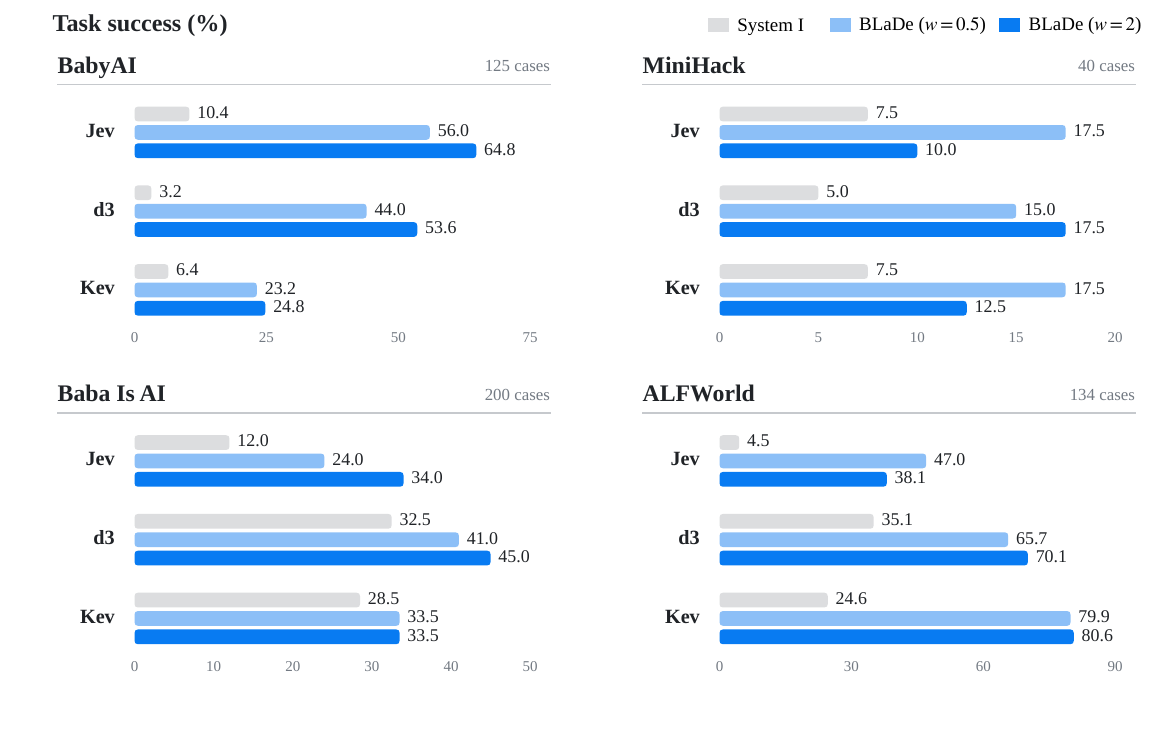}
\else
\input{figures/main_results_bar_plot.tex}
\fi
\caption{\textbf{History and plan guidance across decision models and games.}
Native task-success rates for Jev, d3, and Kev on the four main evaluation environments.
Gray bars show System~I, while light and dark blue bars show \method{} with $w=0.5$ and $w=2$, respectively.
Each panel uses its own labeled horizontal scale.}
\label{fig:main-results-teaser}
\end{figure}

\clearpage

\section{Introduction}
System~I decision models make fast choices by returning probabilities over candidate actions~\citep{typesafe2026jev,vllm2026d3,palmer2026kev27b,chen2026system1arena}, whereas System~II models devote additional computation to deliberative reasoning. In sequential decision-making, individual choices are shaped by past experience and guided by a longer-term goal. Invoking System~II at every step can help maintain this longer-term perspective, but incurs computational overhead. Existing work connects fast and deliberative computation through training, adaptive inference, or modular collaboration~\citep{yu2024distilling,song2026lev,nvidia2025groot,song2025hume}. Modular collaboration pairs periodic System~II planning with fast System~I action selection~\citep{zhang2026jevmobile,ma2026jevstar}. However, new observations and action outcomes can shift the actor’s preferences before the next planning update. Even when the goal remains relevant, how strongly the current plan should steer decisions grounded in recent experience remains unclear.

Incorporating recent history and the planner’s goal into a single actor query allows both to affect the next decision, but leaves the strength of guidance implicit. To make this influence controllable, we treat guidance as an update of strength~$w$ to action preferences grounded in recent interactions~\citep{dpdp,ho2022cfg,li2023contrastive}. From a Bayesian perspective, this update connects past experience with the desired future by reweighting the history-conditioned action prior $\pi_h(a)$ with a future-outcome likelihood. With a factorized future-outcome likelihood, Bayes’ rule expresses its action-dependent contribution as a ratio of conditional action probabilities, allowing it to be computed from the actor’s action probabilities. We instantiate this ratio as $\pi_p(a)/\pi_0(a)$, with the same frozen actor conditioned on the current observation with and without the planner’s target, respectively. Combining this ratio with the history prior provides adjustable goal guidance without separately predicting action outcomes.

We introduce \textbf{B}ayesian \textbf{La}nguage \textbf{De}cision (\textbf{\method{}}), a training-free framework that realizes this rule by dividing the work between a periodic System~II planner and a frozen System~I actor (Figure~\ref{fig:method}). At each scheduled planning update, the planner reads the task, the current observation, and recent interactions, and proposes one desired future state that is held until the next update. Between updates, the actor scores the full action menu three times, once on the current observation alone, once with recent interactions added, and once with the active target added. \method{} then reweights the history-conditioned distribution by the goal ratio raised to the power $w$. Because the history input is renewed after every executed action while the target stays fixed, the actor can respond to new outcomes without waiting for the planner, and $w$ turns the pull of the plan into an explicit setting rather than leaving it fixed inside a single prompt. The composition needs only the option probabilities the actor already returns, so the same procedure applies to API-served and open-weight decision models without fine-tuning or lookahead search.

In experiments with the API-served Jev and the open-weight d3 and Kev, guidance from history and planning helps every actor on all four benchmarks, lifting average success from 14.76\% to 40.38\% at $w=2$. Controls on the three shorter benchmarks show that recent interactions and planner goals each contribute, and that composing them through separate queries slightly exceeds a single joint prompt on average while keeping their balance adjustable. Ablations with d3 further show that four recent interactions capture most of the history benefit and that $w=2$ remains among the best weights, although the optimal weight varies across environments. On ALFWorld, \method{} trades extra inference for 21.8\%--52.5\% fewer environment actions. Our main contributions are summarized below:
\begin{itemize}\setlength{\itemsep}{2pt}
\item \textbf{A Bayesian decision rule.} We formulate the integration of history and planning as Bayesian action selection. Under an outcome factorization, a desired future's likelihood is proportional to a ratio of conditional action probabilities, which a decision model can evaluate without an explicit world model over candidate actions.
\item \textbf{A training-free framework.} \method{} turns this rule into a closed-loop controller that pairs a periodic planner with any frozen System~I actor exposing option probabilities. Three queries to the same actor supply the prior and the ratio, and a single weight $w$ balances past interactions against the planned future.
\item \textbf{Consistent gains with fewer actions.} Without updating the actor, \method{} improves success in every actor--environment pair. On ALFWorld, Kev at $w=2$ rises from 24.6\% to 80.6\% success with 52.5\% fewer environment actions. Controls and ablations examine history, planner goals, joint conditioning, and guidance strength.
\end{itemize}
\section{Related Work}
\paragraph{System~I decision models.}
Jev maps a shared state and typed Boolean, choice, or ordered-score questions directly to option probabilities, providing a decision interface without a generated reasoning trace~\citep{typesafe2026jev}. \citet{deusser2026systemone} examine Jev's decision interface across decision tasks. System1-Arena compares System~I models with generative LLMs and reports that its strongest System~I models attain competitive non-agentic accuracy relative to fast closed LLMs, with better probability estimates and shorter response times~\citep{chen2026system1arena}. Open-weight d3 and Kev provide analogous probability return interfaces with local deployment and native option scoring~\citep{vllm2026d3,palmer2026kev27b}. Narrow Decision Models and MJev-Spatial further explore domain-specific typed decisions and spatial judgments~\citep{antunes2026ndm,qin2026mjevspatial}. These models share a common interface that returns full option distributions, yet System1-Arena's distinction between single-step accuracy and closed-loop success leaves open how they should act over repeated decisions. \method{} builds on this interface, treating any such model as a frozen actor and composing its option distributions to guide closed-loop decisions without training.
\paragraph{Fast--slow collaboration.}
Fast and deliberative computation can be connected through training, adaptive inference, or modular collaboration. \citet{yu2024distilling} distill System~II deliberative outputs into System~I models that answer directly. LEV trains a looped decision model whose inference depth can be adjusted per request~\citep{song2026lev}. GR00T N1 jointly trains vision--language interpretation and diffusion-based action generation~\citep{nvidia2025groot}, while Hume combines low-frequency value-guided action selection with asynchronous reactive control~\citep{song2025hume}. In planner--executor architectures, Jev-Mobile and JEV-Star pair high-level language-model planning with repeated Jev action selection~\citep{zhang2026jevmobile,ma2026jevstar}. Truncated-reasoning experiments study another information path, using Jev to read incomplete LLM reasoning~\citep{wang2026truncated}. \method{} connects a periodic reasoning planner to an already trained, frozen actor. Its focus is how to combine the actor's history-conditioned action preferences with the planner's persistent desired state through separate conditional queries.

\paragraph{History and future guidance.}
Beyond the current observation, agents can draw on two complementary sources of information: past experience and future guidance.
\emph{Past experience.} ReAct, Reflexion, and Voyager carry prior interaction into later decisions through interleaved reasoning, verbal feedback, and reusable skills, respectively~\citep{yao2023react,shinn2023reflexion,wang2024voyager}.
\emph{Future guidance.} Options and goal-conditioned learning formalize temporally extended and goal-directed behavior~\citep{sutton1999options,schaul2015uvfa,andrychowicz2017her}, while language-model planners ground proposals in available actions, affordances, or environment feedback~\citep{huang2022zeroshotplanners,ichter2023saycan,huang2023innermonologue}.
The proposed \method{} supplies one source of each kind directly to a frozen actor: a short buffer of real within-episode interactions and a language-model-proposed desired state.

\paragraph{Distribution composition and guidance.}
Many inference-time control methods compose a base distribution with other distributions or with contrasts between them. \citet{hinton2002poe} combines distributions multiplicatively through products of experts. \citet{liu2021dexperts} introduce DExperts, which reweights a base distribution by an expert-to-anti-expert probability ratio. CFG combines conditional and unconditional denoising scores~\citep{ho2022cfg}, STEVE-1 applies goal-dependent guidance to a Minecraft behavior policy~\citep{lifshitz2023steve1}, and related conditional-versus-reference contrasts appear in offline reinforcement learning~\citep{dpdp}. Context-aware decoding contrasts the same model's predictions with and without additional context~\citep{shi2023cad}, while contrastive decoding (CD) combines expert and amateur likelihoods under a plausibility constraint~\citep{li2023contrastive}. \citet{liu2026loopcd} extend this idea to recurrent computation with LoopCD, contrasting the final and an earlier iteration of the same looped Transformer in logit or hidden state space. Our action policy uses the established expert-ratio form over native option probabilities. It assigns history to the base distribution and the desired state to a separate contrast, making the strength of prospective guidance adjustable while preserving a history-conditioned term in the score.

\section{BLaDe: Bayesian Language Decision}
\subsection{Preliminaries}
After $t$ executed actions, an environment has state $s_t$ and returns a public observation $o_t$. A frozen System~I actor~$\pi$ receives this observation, task instruction $d$, and the complete native action menu $\mathcal A_t$. The observation-only baseline~executes
\begin{equation}
a_t^{\mathrm{System\,I}}=\arg\max_{a\in\mathcal A_t}\pi(a\mid o_t).
\label{eq:flat-policy}
\end{equation}
Executing $a_t$ produces the next observation $o_{t+1}$. Task instructions, public rules, menu descriptions, and mechanical budget fields are shared across input views and suppressed in the policy conditionals below.

\paragraph{Interaction history.}
Let $h_t\equiv h_{t-\ell:t}=((a_i,o_{i+1}))_{i=\max(0,t-\ell)}^{t-1}$ contain the latest $\ell$ real actions and their full native feedback for the actor. The planner reads the same interaction trajectory through its own window, $h_t^{\mathcal G}\equiv h_{t-\ell_{\mathcal G}:t}$, whose length $\ell_{\mathcal G}$ is specified separately from $\ell$.

\paragraph{Planning schedule and desired future.}
Let $T$ be the action budget, $0\leq t_0<T$ the number of history-only warm-up actions, and $1\leq m\leq T-t_0$ the maximum number of planner calls. The fixed refresh counts are
\begin{equation}
\mathcal U_T=\left\{t_0+\left\lfloor\frac{j(T-t_0)}{m}\right\rfloor:j=0,\ldots,m-1\right\}.
\label{eq:schedule}
\end{equation}
Write the ordered elements as $u_k=t_0+\lfloor(k-1)(T-t_0)/m\rfloor$, $k=1,\ldots,m$. The interval endpoint is $\tau_k=u_{k+1}$ for $k<m$, and $\tau_m=T$. At each reached refresh count $u_k$, the System~II planner $\mathcal G$ proposes a desired future-observation description,
\begin{equation}
o'_{\tau_k}\leftarrow\mathcal G(d,o_{u_k},h_{u_k}^{\mathcal G},\mathcal A_{u_k};T-u_k,\tau_k-u_k).
\label{eq:planner-target}
\end{equation}
During interval $k$, the active target is $p_t=o'_{\tau_k}$. Here, $p_t$ describes an intermediate or final environment condition to be pursued within the supplied planning horizon. A condition is considered desirable when the planner judges that pursuing it will help complete task $d$. Each reached refresh produces one target from the trajectory available then; it is held until the next refresh while observations and history continue to update. The planner specifies a desired future, rather than an outcome already observed.

\subsection{Bayesian decision with a planned future}
Fix the shared task, menu, and budget context, and consider a joint probability model with positive conditional probabilities over actions, observations, history, and a future outcome $Y$. Let $p_t$ be its desired value. Conditioning on this value gives the Bayesian action distribution
\begin{equation}
P(a\mid o_t,h_t,Y=p_t)
=\frac{P(a\mid o_t,h_t)P(Y=p_t\mid a,o_t,h_t)}{P(Y=p_t\mid o_t,h_t)}.
\label{eq:bayesian-decision}
\end{equation}
In Equation~\ref{eq:bayesian-decision}, the history-conditioned policy $P(a \mid o_t, h_t)$ serves as the prior, and the likelihood $P(Y = p_t \mid a, o_t, h_t)$ reweights each action by how likely it is to lead to the desired future;
the denominator does not depend on $a$. This likelihood is not directly available, since evaluating it would require predicting the future outcome of every candidate action. We therefore rewrite it in terms of action probabilities, which the actor does provide. To this end, we assume that once the current observation and action are given, the history carries no further information about the future outcome:
\begin{equation}
  P(Y = p_t \mid a, o_t, h_t) = P(Y = p_t \mid a, o_t).
  \label{eq:indep}
\end{equation}
This holds when $o_t$ is a sufficient statistic of the environment state, for example, in a fully observed Markov environment. Under partial observation, Markovian hidden-state dynamics alone do not guarantee it, because $h_t$ may reveal parts of the state that $o_t$ omits; we therefore treat it as a working approximation. Under this assumption, Bayes' rule gives
\begin{equation}
P(Y=p_t\mid a,o_t)
=\frac{P(a\mid o_t,Y=p_t)P(Y=p_t\mid o_t)}{P(a\mid o_t)}.
\label{eq:inverse-likelihood}
\end{equation}
The factor $P(Y=p_t\mid o_t)$ is also action-independent, so
\begin{equation}
P(a\mid o_t,h_t,Y=p_t)
\propto P(a\mid o_t,h_t)\frac{P(a\mid o_t,Y=p_t)}{P(a\mid o_t)}.
\label{eq:bayes-composition}
\end{equation}
Equation~\ref{eq:bayes-composition} thus expresses the posterior through three conditional action distributions, so the update never requires predicting the future outcome of each action, a task that is often computationally intractable.

\paragraph{Powered likelihood.}
To control the influence of the proposed future, we use a powered likelihood \citep{miller2015robustbayesianinferencecoarsening,dpdp} and define the likelihood-weighted distribution as
\begin{equation}
\begin{aligned}
q_w(a\mid o_t,h_t;p_t)
&=\frac{P(a\mid o_t,h_t)P(Y=p_t\mid a,o_t)^w}{C_{t,w}},\\
C_{t,w}&=\sum_{b\in\mathcal A_t}P(b\mid o_t,h_t)P(Y=p_t\mid b,o_t)^w,~~\text{     where } w\geq0.
\end{aligned}
\label{eq:powered-bayes}
\end{equation}
At $w=0$, this retains the history prior; at $w = 1$, it recovers Equation~\ref{eq:bayes-composition}. Substitution of Equation~\ref{eq:inverse-likelihood} yields
\begin{equation}
q_w(a\mid o_t,h_t;p_t)
\propto P(a\mid o_t,h_t)
\left[\frac{P(a\mid o_t,Y=p_t)}{P(a\mid o_t)}\right]^w.
\label{eq:bayes-guidance}
\end{equation}
Increasing $w$ emphasizes the target-induced change relative to the history prior. This derivation motivates the actor queries and composition used by \method{}.

\begin{metalgorithm}{Frozen-actor history--goal control}\label{alg:execution}
\textbf{Input:} actor $\pi$, planner $\mathcal G$, environment $E$, task $d$, action budget $T$, actor history length $\ell$, planner history length $\ell_{\mathcal G}$, weight $w$, refresh steps $\mathcal U_T$, and interval endpoints $\{\tau_k\}$.\\
\textbf{Initialize:} executed count $t=0$, interaction buffer $\mathcal B=\varnothing$, active target $p=\varnothing$.
\setlength{\topsep}{3pt}\setlength{\partopsep}{0pt}
\begin{enumerate}\setlength{\itemsep}{1pt}\setlength{\parskip}{0pt}
\item While $t<T$ and the native episode remains active:
\item\hspace{6pt}Read public observation $o_t$ and complete menu $\mathcal A_t$; read the latest $\ell$ and $\ell_{\mathcal G}$ interactions from $\mathcal B$ as $h_t$ and $h_t^{\mathcal G}$, respectively.
\item\hspace{6pt}If $t=u_k$, obtain $p\leftarrow\mathcal G(d,o_{u_k},h_{u_k}^{\mathcal G},\mathcal A_{u_k};T-u_k,\tau_k-u_k)$, as shown in Equation~\ref{eq:planner-target}.
\item\hspace{6pt}Query $\pih=\pi(\cdot\mid o_t,h_{t-\ell:t})$.
\item\hspace{6pt}If $p=\varnothing$, select $a_t=\arg\max\pih$; otherwise query $\piz,\pip$ and, with $p_t=p$, select the action as shown in Equation~\ref{eq:score}:
\begingroup
\setlength{\abovedisplayskip}{2pt}\setlength{\belowdisplayskip}{2pt}%
\setlength{\abovedisplayshortskip}{2pt}\setlength{\belowdisplayshortskip}{2pt}%
\begin{equation*}
S_w(a)=\log\pih(a)+w\bigl[\log\pip(a)-\log\piz(a)\bigr],\quad
 a_t=\arg\max_{a\in\mathcal A_t}S_w(a).
\end{equation*}
\endgroup
\item\hspace{6pt}Execute $a_t$ once in $E$, receive $o_{t+1}$, append $(a_t,o_{t+1})$ to $\mathcal B$, and increment $t$.
\item End while; return the native outcome and interaction trajectory.
\end{enumerate}
\end{metalgorithm}

\subsection{From Bayes' rule to actor queries}
\paragraph{One actor, three conditional views.}
We instantiate the Bayesian factors through three queries to the same frozen actor:
\begin{equation}
\begin{aligned}
\piz(a)&=\pi(a\mid o_t), &\quad&\text{current observation},\\
\pih(a)&=\pi(a\mid o_t,h_t), &&\text{observation and history},\\
\pip(a)&=\pi(a\mid o_t,p_t), &&\text{observation and desired future}.
\end{aligned}
\label{eq:branches}
\end{equation}
All views use shared weights and the same complete menu. The history query receives real interaction records; the future query receives the planner's target, which can itself reflect those records. These readouts are the operational counterparts of the conditional probabilities in the derivation. Their composition is
\begin{equation}
\widetilde\pi_w(a)=\frac{1}{Z_t}\pih(a)
\left(\frac{\pip(a)}{\piz(a)}\right)^w,
\qquad Z_t=\sum_{b\in\mathcal A_t}\pih(b)
\left(\frac{\pip(b)}{\piz(b)}\right)^w.
\label{eq:product}
\end{equation}
For numerical stability, we take logarithms and execute the mode:
\begin{equation}
\begin{aligned}
S_w(a)&=\log\pih(a)+w\bigl[\log\pip(a)-\log\piz(a)\bigr],\\
a_t&=\arg\max_{a\in\mathcal A_t}S_w(a).
\end{aligned}
\label{eq:score}
\end{equation}
Returned distributions are normalized and positively smoothed before scoring; exact ties follow the frozen menu order. Numerical settings appear in the experimental setup.

If the actor readouts coincide with the joint-model conditionals, this implements Equation~\ref{eq:bayes-guidance}. For any positive readouts, the same composition also has an exact KL-regularized interpretation; the derivation and target-contrast interpretation are provided in Appendix~\ref{app:interpretation}.

\paragraph{Closed-loop execution.}
Before the first target refresh, the actor follows $\pih$. Thereafter, Equation~\ref{eq:score} combines the three views for each decision. Native feedback updates $o_t$ and $h_t$ after every real action, while the target stays fixed within its planning interval. At the next scheduled refresh, the planner replaces it using the updated trajectory. Algorithm~\ref{alg:execution} and Figure~\ref{fig:method} summarize this control loop.

\begin{figure}[!htbp]
\centering
\input{figures/method_figure.tex}
\end{figure}

\section{Experiments}
\subsection{Experimental setup}
\paragraph{Actors and planner.}
We evaluate Jev (configured as \nolinkurl{jev-1.13.0}), d3 (\nolinkurl{vllm-sr/d3}, 27B), and Kev (\nolinkurl{jaredpalmer/kev-27b}). The latter two run locally in BF16 with their native readouts. No actor is fine-tuned for this study. The planner is configured as \texttt{gpt-6-sol} with medium reasoning effort and a 4,096-token output budget, returning one JSON \texttt{goal} field. Exact readout, provider, and interface details are provided in Appendix~\ref{app:implementation}.

\begingroup
\setlength{\emergencystretch}{1.5em}
\paragraph{Environments and cases.}
The main evaluation comprises BabyAI \citep{chevalierboisvert2019babyai}, MiniHack \citep{samvelyan2021minihack}, Baba Is AI~\citep{cloos2024baba}, and ALFWorld~\citep{shridhar2021alfworld}. The four main environments contribute 125, 40, 200, and 134 cases, respectively. Each actor and condition is evaluated on the same 499 cases; three actors and three conditions give 4,491 scheduled episodes. The three short environments use BALROG-derived native text adapters~\citep{paglieri2025balrog}; ALFWorld uses its textual setting and official full current admissible-action menu for all conditions. Case counts, action budgets, and native menus are summarized in Table~\ref{tab:protocol}; the fixed case bank and interface details are provided in Appendix~\ref{app:implementation}.

\par
\endgroup

\paragraph{Conditions.}
\textbf{System~I} selects $\arg\max\piz$ from the first action using only the current observation and shared task context, without past interactions or planner goals. The two \textbf{\method{}} conditions additionally use up to the four most recent action--observation pairs ($\ell=\ell_{\mathcal G}=4$) and a periodically refreshed planner goal, with respective fixed guidance weights $w=0.5$ and $w=2$. Each condition builds its history and goals from its own interaction trajectory. Our main experiments evaluate the overall effectiveness of \method{}, integrating history and planning guidance for action selection.

\paragraph{Planning schedule and numerical settings.}
In the four main environments, we instantiate Equation~\ref{eq:schedule} with an eight-action history-only warm-up ($t_0=8$) and at most four planner calls per episode ($m=4$). For ALFWorld, $T=50$ gives refresh counts $\{8,18,29,39\}$: the first goal is generated after eight executed actions, before selecting action~9. A goal is retained until the next fixed refresh or the end of the episode. For the composed score, each returned branch distribution is normalized and smoothed as $\bar\pi(a)=(\pi(a)+\epsilon/|\mathcal A_t|)/(1+\epsilon)$, with $\epsilon=10^{-6}$, before evaluating the action score. Exact ties use the frozen menu order.

\paragraph{Metrics.}
We report native task-success rate: the proportion of scheduled evaluation episodes that satisfy the environment's task objective within the action budget. Task success is defined for each benchmark as follows. \textbf{BabyAI:} completing the instruction, including navigation, pickup, opening, placement, or sequential tasks. \textbf{MiniHack:} reaching the native \texttt{TASK\_SUCCESSFUL} terminal status. \textbf{Baba Is AI:} satisfying the native level-win condition under the active rules. \textbf{ALFWorld:} completing the full native household-task goal. Efficiency is measured by mean executed actions per completed episode. Token usage and API cost are reported separately.

\subsection{Main Results}
Table~\ref{tab:main} shows that both guidance weights improve success over the corresponding actor's System~I baseline across all three actors and all four environments, giving 24 positive comparisons. With $w=2$, success averaged equally across actors and environments rises from 14.76\% to 40.38\%. For example, Jev on BabyAI rises from 10.4\% to 64.8\%, d3 on Baba Is AI from 32.5\% to 45.0\%, and Kev on ALFWorld from 24.6\% to 80.6\%. These results show that combining history and goal guidance substantially improves frozen decision models across diverse interactive environments.

\begin{table}[t]
\centering\small
\caption{\textbf{Success across actors and environments.} Cells show success rates (\%) and counts. System~I uses the current observation; \method{} adds recent interactions and planner goals. Bold marks the highest rate per actor and environment, including ties.}
\label{tab:main}
\setlength{\tabcolsep}{4pt}
\renewcommand{\arraystretch}{1.03}
\begin{tabular*}{0.8\linewidth}{@{\extracolsep{\fill}}llrrrr@{}}
\toprule
Actor & Method & \makecell{BabyAI\\$N=125$} & \makecell{MiniHack\\$N=40$} & \makecell{Baba Is AI\\$N=200$} & \makecell{ALFWorld\\$N=134$} \\
\midrule
\multirow{3}{*}{Jev} & System~I & 10.4 (13) & 7.5 (3) & 12.0 (24) & 4.5 (6) \\
 & \method{}, $w=0.5$ & 56.0 (70) & \textbf{17.5} (7) & 24.0 (48) & \textbf{47.0} (63) \\
 & \method{}, $w=2$ & \textbf{64.8} (81) & 10.0 (4) & \textbf{34.0} (68) & 38.1 (51) \\
\midrule
\multirow{3}{*}{d3} & System~I & 3.2 (4) & 5.0 (2) & 32.5 (65) & 35.1 (47) \\
 & \method{}, $w=0.5$ & 44.0 (55) & 15.0 (6) & 41.0 (82) & 65.7 (88) \\
 & \method{}, $w=2$ & \textbf{53.6} (67) & \textbf{17.5} (7) & \textbf{45.0} (90) & \textbf{70.1} (94) \\
\midrule
\multirow{3}{*}{Kev} & System~I & 6.4 (8) & 7.5 (3) & 28.5 (57) & 24.6 (33) \\
 & \method{}, $w=0.5$ & 23.2 (29) & \textbf{17.5} (7) & \textbf{33.5} (67) & 79.9 (107) \\
 & \method{}, $w=2$ & \textbf{24.8} (31) & 12.5 (5) & \textbf{33.5} (67) & \textbf{80.6} (108) \\
\bottomrule
\end{tabular*}
\vspace{3mm}
\end{table}

Table~\ref{tab:alf} reports repairs and damage rather than only aggregate accuracy. Every guided condition repairs substantially more baseline failures than it damages baseline successes. Kev with $w=2$ repairs 76 cases and damages one, producing a paired gain of 55.97 percentage points. A sensitivity analysis on the 132 cases completed by every condition preserves the broad pattern (Appendix~\ref{app:statistics}). The primary analysis continues to use all 134 cases.

\subsection{Ablation Study}\label{sec:ablation}
\subsubsection{History and goal conditioning}
We compare two controls with Jev and d3 on BabyAI, MiniHack, and Baba Is AI. \textbf{History} selects actions using the current observation and the four most recent interactions, without planner goals. \textbf{Joint} supplies the observation, the same history window, and the active goal in a single actor query. Joint uses the same warm-up and fixed goal-refresh schedule as \method{}. Figure~\ref{fig:short-ablation-five-arms} and Table~\ref{tab:short-ablation} compare these controls with System~I and the two guidance weights; implementation details appear in Appendix~\ref{app:ablation-pairs}.

\begin{figure}[!htbp]
\centering
\setlength{\parskip}{0pt}
\newsavebox{\JanusPairedTableBox}
\newsavebox{\JanusPairedRadarBox}
\newlength{\JanusPairedContentHeight}
\sbox{\JanusPairedTableBox}{%
\begingroup
\small
\setlength{\tabcolsep}{3pt}
\renewcommand{\arraystretch}{1.50}
\setlength{\aboverulesep}{1.5pt}
\setlength{\belowrulesep}{2pt}
\begin{tabular}{@{}llrrr@{\hspace{6pt}}r@{}}
\toprule
Actor & Method & Successes & Repair & Damage & $\Delta$ (pp) \\
\midrule
Jev & System~I    & 6   & -- & -- & --     \\
    & $w=0.5$ & 63  & 59 & 2  & +42.54 \\
    & $w=2$   & 51  & 47 & 2  & +33.58 \\
\midrule
d3  & System~I    & 47  & -- & -- & --     \\
    & $w=0.5$ & 88  & 45 & 4  & +30.60 \\
    & $w=2$   & 94  & 48 & 1  & +35.07 \\
\midrule
Kev & System~I    & 33  & -- & -- & --     \\
    & $w=0.5$ & 107 & 75 & 1  & +55.22 \\
    & $w=2$   & 108 & 76 & 1  & +55.97 \\
\bottomrule
\end{tabular}
\endgroup}
\setlength{\JanusPairedContentHeight}{%
  \dimexpr\ht\JanusPairedTableBox+\dp\JanusPairedTableBox\relax}
\sbox{\JanusPairedRadarBox}{%
  \resizebox*{!}{\JanusPairedContentHeight}{
\begingroup
\begin{tikzpicture}[x=1cm,y=1cm,font=\rmfamily\small]
\path[use as bounding box] (-4.55,-3.50) rectangle (4.55,4.30);
\definecolor{radflat}{HTML}{3875C6}
\definecolor{radhistory}{HTML}{4DA564}
\definecolor{radjoint}{HTML}{9270AF}
\definecolor{radhalf}{HTML}{E5A23B}
\definecolor{radtwo}{HTML}{D75B64}
\definecolor{radGrid}{HTML}{D2D6DC}
\definecolor{radSpoke}{HTML}{BEC4CE}
\definecolor{radText}{HTML}{263445}
\definecolor{radTick}{HTML}{818A99}
\begin{pgfonlayer}{background}
\path[fill=radflat,fill opacity=0.12,draw=none] (0.000000,0.364000) -- (0.096995,0.056000) -- (0.227332,-0.131250) -- (0.000000,-0.175000) -- (-0.363731,-0.210000) -- (-0.985104,0.568750) -- cycle;
\path[fill=radhistory,fill opacity=0.12,draw=none] (0.000000,1.624000) -- (1.333679,0.770000) -- (0.227332,-0.131250) -- (0.000000,-0.437500) -- (-0.530441,-0.306250) -- (-1.151814,0.665000) -- cycle;
\path[fill=radjoint,fill opacity=0.12,draw=none] (0.000000,1.848000) -- (1.382177,0.798000) -- (0.454663,-0.262500) -- (0.000000,-0.700000) -- (-1.197280,-0.691250) -- (-1.318524,0.761250) -- cycle;
\path[fill=radhalf,fill opacity=0.12,draw=none] (0.000000,1.960000) -- (1.333679,0.770000) -- (0.530441,-0.306250) -- (0.000000,-0.525000) -- (-0.727461,-0.420000) -- (-1.242746,0.717500) -- cycle;
\path[fill=radtwo,fill opacity=0.12,draw=none] (0.000000,2.268000) -- (1.624664,0.938000) -- (0.303109,-0.175000) -- (0.000000,-0.612500) -- (-1.030570,-0.595000) -- (-1.363990,0.787500) -- cycle;
\end{pgfonlayer}
\draw[radGrid,line width=0.55pt] (0,0) circle (0.350000);
\draw[radGrid,line width=0.55pt] (0,0) circle (0.700000);
\draw[radGrid,line width=0.55pt] (0,0) circle (1.050000);
\draw[radGrid,line width=0.55pt] (0,0) circle (1.400000);
\draw[radGrid,line width=0.55pt] (0,0) circle (1.750000);
\draw[radGrid,line width=0.55pt] (0,0) circle (2.100000);
\draw[radGrid,line width=0.75pt] (0,0) circle (2.450000);
\draw[radSpoke,line width=0.6pt] (0,0) -- (0.000000,2.450000);
\node[align=center,text=radText,anchor=south,inner sep=0pt,font=\rmfamily\fontsize{9}{10}\selectfont] at (0.000000,2.810000) {BabyAI\\Jev};
\draw[radSpoke,line width=0.6pt] (0,0) -- (2.121762,1.225000);
\node[align=center,text=radText,anchor=west,inner sep=0pt,font=\rmfamily\fontsize{9}{10}\selectfont] at (2.433531,1.405000) {BabyAI\\d3};
\draw[radSpoke,line width=0.6pt] (0,0) -- (2.121762,-1.225000);
\node[align=center,text=radText,anchor=west,inner sep=0pt,font=\rmfamily\fontsize{9}{10}\selectfont] at (2.433531,-1.405000) {MiniHack\\Jev};
\draw[radSpoke,line width=0.6pt] (0,0) -- (0.000000,-2.450000);
\node[align=center,text=radText,anchor=north,inner sep=0pt,font=\rmfamily\fontsize{9}{10}\selectfont] at (0.000000,-2.810000) {MiniHack\\d3};
\draw[radSpoke,line width=0.6pt] (0,0) -- (-2.121762,-1.225000);
\node[align=center,text=radText,anchor=east,inner sep=0pt,font=\rmfamily\fontsize{9}{10}\selectfont] at (-2.433531,-1.405000) {Baba Is AI\\Jev};
\draw[radSpoke,line width=0.6pt] (0,0) -- (-2.121762,1.225000);
\node[align=center,text=radText,anchor=east,inner sep=0pt,font=\rmfamily\fontsize{9}{10}\selectfont] at (-2.433531,1.405000) {Baba Is AI\\d3};
\draw[radflat,line width=1.1pt,line join=round] (0.000000,0.364000) -- (0.096995,0.056000) -- (0.227332,-0.131250) -- (0.000000,-0.175000) -- (-0.363731,-0.210000) -- (-0.985104,0.568750) -- cycle;
\filldraw[fill=radflat,draw=white,line width=0.3pt] (0.000000,0.364000) circle (0.50mm);
\filldraw[fill=radflat,draw=white,line width=0.3pt] (0.096995,0.056000) circle (0.50mm);
\filldraw[fill=radflat,draw=white,line width=0.3pt] (0.227332,-0.131250) circle (0.50mm);
\filldraw[fill=radflat,draw=white,line width=0.3pt] (0.000000,-0.175000) circle (0.50mm);
\filldraw[fill=radflat,draw=white,line width=0.3pt] (-0.363731,-0.210000) circle (0.50mm);
\filldraw[fill=radflat,draw=white,line width=0.3pt] (-0.985104,0.568750) circle (0.50mm);
\draw[radhistory,line width=1.1pt,line join=round,dash pattern=on 3pt off 2pt] (0.000000,1.624000) -- (1.333679,0.770000) -- (0.227332,-0.131250) -- (0.000000,-0.437500) -- (-0.530441,-0.306250) -- (-1.151814,0.665000) -- cycle;
\filldraw[fill=radhistory,draw=white,line width=0.3pt] (0.000000,1.624000) circle (0.50mm);
\filldraw[fill=radhistory,draw=white,line width=0.3pt] (1.333679,0.770000) circle (0.50mm);
\filldraw[fill=radhistory,draw=white,line width=0.3pt] (0.227332,-0.131250) circle (0.50mm);
\filldraw[fill=radhistory,draw=white,line width=0.3pt] (0.000000,-0.437500) circle (0.50mm);
\filldraw[fill=radhistory,draw=white,line width=0.3pt] (-0.530441,-0.306250) circle (0.50mm);
\filldraw[fill=radhistory,draw=white,line width=0.3pt] (-1.151814,0.665000) circle (0.50mm);
\draw[radjoint,line width=1.1pt,line join=round,dash pattern=on 3pt off 1.4pt on 0.7pt off 1.4pt] (0.000000,1.848000) -- (1.382177,0.798000) -- (0.454663,-0.262500) -- (0.000000,-0.700000) -- (-1.197280,-0.691250) -- (-1.318524,0.761250) -- cycle;
\filldraw[fill=radjoint,draw=white,line width=0.3pt] (0.000000,1.848000) circle (0.50mm);
\filldraw[fill=radjoint,draw=white,line width=0.3pt] (1.382177,0.798000) circle (0.50mm);
\filldraw[fill=radjoint,draw=white,line width=0.3pt] (0.454663,-0.262500) circle (0.50mm);
\filldraw[fill=radjoint,draw=white,line width=0.3pt] (0.000000,-0.700000) circle (0.50mm);
\filldraw[fill=radjoint,draw=white,line width=0.3pt] (-1.197280,-0.691250) circle (0.50mm);
\filldraw[fill=radjoint,draw=white,line width=0.3pt] (-1.318524,0.761250) circle (0.50mm);
\draw[radhalf,line width=1.45pt,line join=round] (0.000000,1.960000) -- (1.333679,0.770000) -- (0.530441,-0.306250) -- (0.000000,-0.525000) -- (-0.727461,-0.420000) -- (-1.242746,0.717500) -- cycle;
\filldraw[fill=radhalf,draw=white,line width=0.3pt] (0.000000,1.960000) circle (0.57mm);
\filldraw[fill=radhalf,draw=white,line width=0.3pt] (1.333679,0.770000) circle (0.57mm);
\filldraw[fill=radhalf,draw=white,line width=0.3pt] (0.530441,-0.306250) circle (0.57mm);
\filldraw[fill=radhalf,draw=white,line width=0.3pt] (0.000000,-0.525000) circle (0.57mm);
\filldraw[fill=radhalf,draw=white,line width=0.3pt] (-0.727461,-0.420000) circle (0.57mm);
\filldraw[fill=radhalf,draw=white,line width=0.3pt] (-1.242746,0.717500) circle (0.57mm);
\draw[radtwo,line width=1.45pt,line join=round] (0.000000,2.268000) -- (1.624664,0.938000) -- (0.303109,-0.175000) -- (0.000000,-0.612500) -- (-1.030570,-0.595000) -- (-1.363990,0.787500) -- cycle;
\filldraw[fill=radtwo,draw=white,line width=0.3pt] (0.000000,2.268000) circle (0.57mm);
\filldraw[fill=radtwo,draw=white,line width=0.3pt] (1.624664,0.938000) circle (0.57mm);
\filldraw[fill=radtwo,draw=white,line width=0.3pt] (0.303109,-0.175000) circle (0.57mm);
\filldraw[fill=radtwo,draw=white,line width=0.3pt] (0.000000,-0.612500) circle (0.57mm);
\filldraw[fill=radtwo,draw=white,line width=0.3pt] (-1.030570,-0.595000) circle (0.57mm);
\filldraw[fill=radtwo,draw=white,line width=0.3pt] (-1.363990,0.787500) circle (0.57mm);
\node[text=radTick,fill=white,inner sep=0.3pt,font=\rmfamily\fontsize{7.2}{8}\selectfont] at (0.370000,0.000000) {10};
\node[text=radTick,fill=white,inner sep=0.3pt,font=\rmfamily\fontsize{7.2}{8}\selectfont] at (0.720000,0.000000) {20};
\node[text=radTick,fill=white,inner sep=0.3pt,font=\rmfamily\fontsize{7.2}{8}\selectfont] at (1.070000,0.000000) {30};
\node[text=radTick,fill=white,inner sep=0.3pt,font=\rmfamily\fontsize{7.2}{8}\selectfont] at (1.420000,0.000000) {40};
\node[text=radTick,fill=white,inner sep=0.3pt,font=\rmfamily\fontsize{7.2}{8}\selectfont] at (1.770000,0.000000) {50};
\node[text=radTick,fill=white,inner sep=0.3pt,font=\rmfamily\fontsize{7.2}{8}\selectfont] at (2.120000,0.000000) {60};
\node[text=radTick,fill=white,inner sep=0.3pt,font=\rmfamily\fontsize{7.2}{8}\selectfont] at (2.470000,0.000000) {70};
\draw[radflat,line width=1.1pt] (-3.600000,3.980000) -- (-3.220000,3.980000);
\node[anchor=west,inner sep=0pt,font=\rmfamily\fontsize{8.6}{9.6}\selectfont] at (-3.130000,3.980000) {System~I};
\draw[radhistory,line width=1.1pt,dash pattern=on 3pt off 2pt] (-1.350000,3.980000) -- (-0.970000,3.980000);
\node[anchor=west,inner sep=0pt,font=\rmfamily\fontsize{8.6}{9.6}\selectfont] at (-0.880000,3.980000) {History};
\draw[radjoint,line width=1.1pt,dash pattern=on 3pt off 1.4pt on 0.7pt off 1.4pt] (1.200000,3.980000) -- (1.580000,3.980000);
\node[anchor=west,inner sep=0pt,font=\rmfamily\fontsize{8.6}{9.6}\selectfont] at (1.670000,3.980000) {Joint};
\draw[radhalf,line width=1.45pt] (-2.100000,3.530000) -- (-1.720000,3.530000);
\node[anchor=west,inner sep=0pt,font=\rmfamily\fontsize{8.6}{9.6}\selectfont] at (-1.630000,3.530000) {\method{}, $w=0.5$};
\draw[radtwo,line width=1.45pt] (0.900000,3.530000) -- (1.280000,3.530000);
\node[anchor=west,inner sep=0pt,font=\rmfamily\fontsize{8.6}{9.6}\selectfont] at (1.370000,3.530000) {\method{}, $w=2$};
\end{tikzpicture}
\endgroup}}
\ifdim\dimexpr\ht\JanusPairedRadarBox+\dp\JanusPairedRadarBox-\JanusPairedContentHeight\relax>0.1pt
  \PackageError{Janus}{Paired radar is taller than the table}{Both content boxes must have equal total height.}
\fi
\ifdim\dimexpr\JanusPairedContentHeight-\ht\JanusPairedRadarBox-\dp\JanusPairedRadarBox\relax>0.1pt
  \PackageError{Janus}{Paired radar is shorter than the table}{Both content boxes must have equal total height.}
\fi
\ifdim\wd\JanusPairedRadarBox>0.49\linewidth
  \PackageError{Janus}{Paired radar exceeds its column}{Reduce the shared content height.}
\fi
\noindent
\begin{minipage}[t]{0.48\linewidth}
\vspace{0pt}
\captionsetup{font=small,skip=4pt}
\captionof{table}{\textbf{Paired ALFWorld gains.} Successes are counts over 134 cases; rows labeled by $w$ use \method{}. Repair counts corrected System~I failures and damage counts lost System~I successes; $\Delta$ is the resulting change.}
\label{tab:alf}
\centering
\makebox[\linewidth][c]{%
  \raisebox{\dp\JanusPairedTableBox}{\usebox{\JanusPairedTableBox}}}
\end{minipage}\hfill
\begin{minipage}[t]{0.49\linewidth}
\vspace{0pt}
\centering
\makebox[\linewidth][c]{%
  \raisebox{\dp\JanusPairedRadarBox}{\usebox{\JanusPairedRadarBox}}}
\par\vspace{2pt}
\captionsetup{font=small,skip=0pt}
\captionof{figure}{\textbf{History and goal guidance.} Success rates (\%) on six actor--environment pairs, with a common 0--70 scale. Exact rates and counts appear in Table~\ref{tab:short-ablation}.}
\label{fig:short-ablation-five-arms}
\end{minipage}
\vspace{3mm}
\end{figure}

History improves success over System~I in five of six actor--environment comparisons and ties it for Jev on MiniHack. Adding planner goals through Joint improves observed success over History in all six comparisons. On Baba Is AI, Jev increases from 17.5\% to 39.5\% and d3 from 38.0\% to 43.5\%, demonstrating the value of prospective goals in addition to recent experience.

Separate probability composition provides an additional way to control this guidance. With $w=2$, \method{} reaches 64.8\% on BabyAI for Jev and 53.6\% for d3, compared with 52.8\% and 45.6\% for Joint. Averaged equally across both actors and all three environments, success is 37.48\%, versus 36.07\% for Joint. The advantage depends on the actor and environment: Joint performs better on MiniHack for both actors and on Baba Is AI for Jev.

\subsubsection{Actor history length}
\begin{figure}[!htbp]
\vspace{1mm}
\centering
\newcommand{\cnt}[1]{\,{\scriptsize(#1)}}
\begin{minipage}[t]{0.47\linewidth}
\vspace{0pt}
\captionsetup{font=small,skip=4pt,justification=raggedright,singlelinecheck=false}
\captionof{table}{\textbf{History, joint conditioning, and composition.} Success rates (\%) with counts in parentheses, over 125 BabyAI, 40 MiniHack, and 200 Baba Is AI cases. History uses the four most recent interactions; Joint adds the active planner goal in the same actor query; rows labeled by $w$ use \method{}. Bold marks the highest rate per column, including ties.}
\label{tab:short-ablation}
\centering\footnotesize
\setlength{\tabcolsep}{2.4pt}
\renewcommand{\arraystretch}{1.15}
\begin{tabular}{@{}lrrrrrr@{}}
\toprule
 & \multicolumn{2}{c}{BabyAI} & \multicolumn{2}{c}{MiniHack} & \multicolumn{2}{c}{Baba Is AI} \\
\cmidrule(lr){2-3}\cmidrule(lr){4-5}\cmidrule(l){6-7}
Method & Jev & d3 & Jev & d3 & Jev & d3 \\
\midrule
System~I & 10.4\cnt{13} & 3.2\cnt{4} & 7.5\cnt{3} & 5.0\cnt{2} & 12.0\cnt{24} & 32.5\cnt{65} \\
History & 46.4\cnt{58} & 44.0\cnt{55} & 7.5\cnt{3} & 12.5\cnt{5} & 17.5\cnt{35} & 38.0\cnt{76} \\
Joint & 52.8\cnt{66} & 45.6\cnt{57} & 15.0\cnt{6} & \textbf{20.0}\cnt{8} & \textbf{39.5}\cnt{79} & 43.5\cnt{87} \\
\midrule
$w=0.5$ & 56.0\cnt{70} & 44.0\cnt{55} & \textbf{17.5}\cnt{7} & 15.0\cnt{6} & 24.0\cnt{48} & 41.0\cnt{82} \\
$w=2$ & \textbf{64.8}\cnt{81} & \textbf{53.6}\cnt{67} & 10.0\cnt{4} & 17.5\cnt{7} & 34.0\cnt{68} & \textbf{45.0}\cnt{90} \\
\bottomrule
\end{tabular}
\end{minipage}\hfill
\begin{minipage}[t]{0.50\linewidth}
\vspace{0pt}
\centering
\IfFileExists{figures/history_length_babyai.pdf}{%
\includegraphics[width=\linewidth]{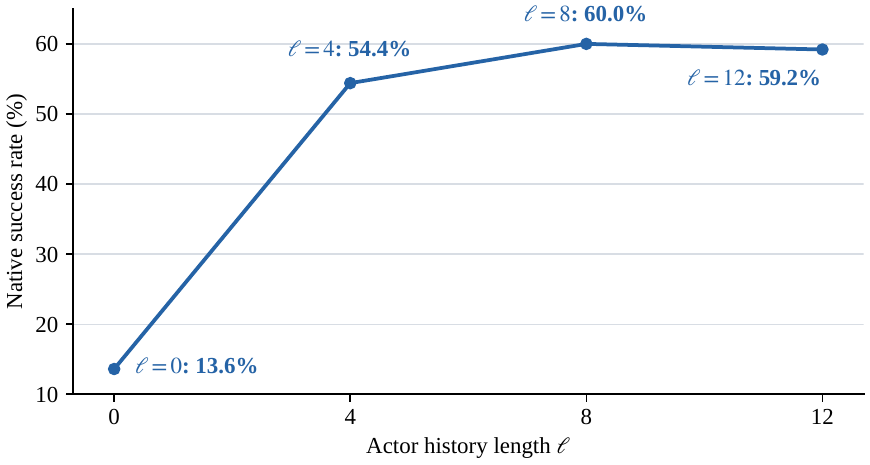}%
}{\resizebox{\linewidth}{!}{\resizebox{0.76\linewidth}{!}{%
\begin{tikzpicture}[x=1cm,y=1cm,font=\rmfamily\fontsize{10}{11}\selectfont]
\path[use as bounding box] (-1,-.95) rectangle (9.1,4.6);
\draw[gray!25,line width=.4pt] (0,0.000000) -- (8.6,0.000000);
\node[anchor=east] at (-.12,0.000000) {10};
\draw[gray!25,line width=.4pt] (0,0.745455) -- (8.6,0.745455);
\node[anchor=east] at (-.12,0.745455) {20};
\draw[gray!25,line width=.4pt] (0,1.490909) -- (8.6,1.490909);
\node[anchor=east] at (-.12,1.490909) {30};
\draw[gray!25,line width=.4pt] (0,2.236364) -- (8.6,2.236364);
\node[anchor=east] at (-.12,2.236364) {40};
\draw[gray!25,line width=.4pt] (0,2.981818) -- (8.6,2.981818);
\node[anchor=east] at (-.12,2.981818) {50};
\draw[gray!25,line width=.4pt] (0,3.727273) -- (8.6,3.727273);
\node[anchor=east] at (-.12,3.727273) {60};
\draw[gray!70] (0,4.1) -- (0,0) -- (8.6,0);
\draw[metablue,line width=1.3pt] (0.000000,0.268364) -- (2.866667,3.309818) -- (5.733333,3.727273) -- (8.600000,3.667636);
\fill[metablue] (0.000000,0.268364) circle(.055);
\node[anchor=west,text=metablue,font=\rmfamily\bfseries\fontsize{11}{12}\selectfont] at (0.160000,0.268364) {$\ell=0$: 13.6\%};
\node[anchor=north] at (0.000000,-.12) {0};
\fill[metablue] (2.866667,3.309818) circle(.055);
\node[anchor=south,text=metablue,font=\rmfamily\bfseries\fontsize{11}{12}\selectfont] at (2.866667,3.469818) {$\ell=4$: 54.4\%};
\node[anchor=north] at (2.866667,-.12) {4};
\fill[metablue] (5.733333,3.727273) circle(.055);
\node[anchor=south,text=metablue,font=\rmfamily\bfseries\fontsize{11}{12}\selectfont] at (5.733333,3.857273) {$\ell=8$: 60.0\%};
\node[anchor=north] at (5.733333,-.12) {8};
\fill[metablue] (8.600000,3.667636) circle(.055);
\node[anchor=north east,text=metablue,font=\rmfamily\bfseries\fontsize{11}{12}\selectfont] at (8.600000,3.507636) {$\ell=12$: 59.2\%};
\node[anchor=north] at (8.600000,-.12) {12};
\node at (4.3,-.72) {Actor history length $\ell$};
\node[rotate=90] at (-.78,2.05) {Native success rate (\%)};
\end{tikzpicture}%
}}}
\par\vspace{4pt}
\captionsetup{font=small,skip=0pt}
\captionof{figure}{\textbf{Actor history length at fixed $w=2$.} d3 native task-success rates in an independent experiment on all 125 BabyAI cases at $\ell=0,4,8,12$ (500 completed episodes). Only the actor's history window changes; the planner always receives the four most recent interactions ($\ell_{\mathcal G}=4$) under the same fixed update schedule (Appendix~\ref{app:full-history-ablation}).}
\label{fig:history-length-babyai}
\end{minipage}
\end{figure}

In an independent history-length experiment, we vary d3's actor history over $\ell\in\{0,4,8,12\}$ recent interactions at fixed $w=2$ on all 125 BabyAI cases, giving 500 episodes. To isolate the actor's memory window, the planner always receives the four most recent interactions ($\ell_{\mathcal G}=4$) under the same model and update schedule; each condition follows its own trajectory. Removing the actor's history therefore retains planner goals and differs from System~I. Figure~\ref{fig:history-length-babyai} shows that success rises from 13.6\% without actor history to 54.4\% with four interactions and 60.0\% with eight; extending the window to twelve gives no further improvement (59.2\%). Four interactions capture most of the observed history benefit with a shorter input context. We retain this compact window in the main configuration, while the additional gain at eight interactions motivates adapting memory length to the task. The full 125-case protocol and the independently executed $\ell=4$ reference appear in Appendix~\ref{app:full-history-ablation}. Separate small-cohort history and planner-variant comparisons are reported in Appendices~\ref{app:local-history-exploration} and~\ref{app:planner-variants}.

\subsubsection{Guidance strength}\label{sec:weight-ablation}
\begin{figure}[!tbp]
\centering
\IfFileExists{figures/guidance_weight_by_dataset.pdf}{%
\includegraphics[width=\linewidth]{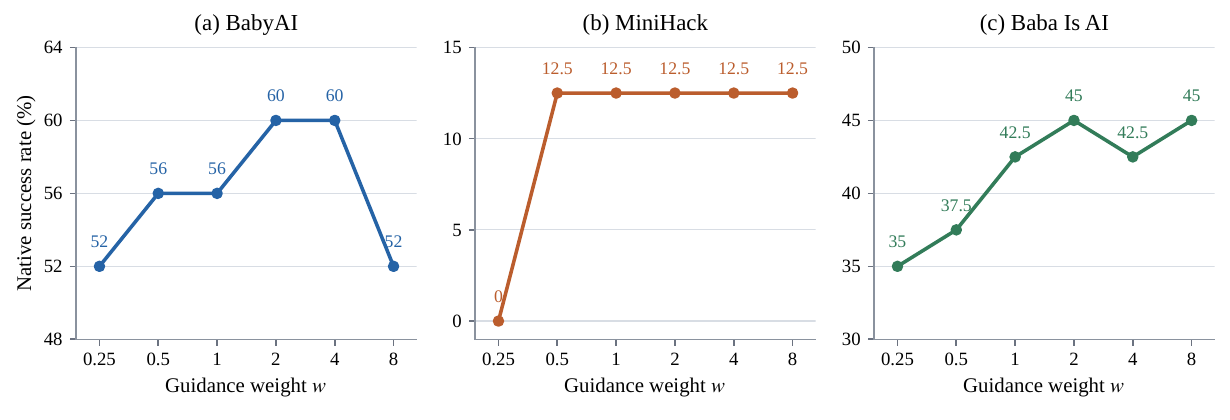}%
}{
\resizebox{\linewidth}{!}{%
\begin{tikzpicture}[x=1cm,y=1cm,font=\rmfamily\fontsize{9}{10}\selectfont]
\path[use as bounding box] (-0.9,-1.0) rectangle (15.4,4.25);
\definecolor{wcolor0}{HTML}{2563A6}
\draw[gray!25,line width=0.4pt] (0.00000,0.00000) -- (4.20000,0.00000);
\node[anchor=east,font=\rmfamily\fontsize{8.5}{9.5}\selectfont] at (-0.12000,0.00000) {48};
\draw[gray!25,line width=0.4pt] (0.00000,0.81250) -- (4.20000,0.81250);
\node[anchor=east,font=\rmfamily\fontsize{8.5}{9.5}\selectfont] at (-0.12000,0.81250) {52};
\draw[gray!25,line width=0.4pt] (0.00000,1.62500) -- (4.20000,1.62500);
\node[anchor=east,font=\rmfamily\fontsize{8.5}{9.5}\selectfont] at (-0.12000,1.62500) {56};
\draw[gray!25,line width=0.4pt] (0.00000,2.43750) -- (4.20000,2.43750);
\node[anchor=east,font=\rmfamily\fontsize{8.5}{9.5}\selectfont] at (-0.12000,2.43750) {60};
\draw[gray!25,line width=0.4pt] (0.00000,3.25000) -- (4.20000,3.25000);
\node[anchor=east,font=\rmfamily\fontsize{8.5}{9.5}\selectfont] at (-0.12000,3.25000) {64};
\draw[gray!70,line width=.5pt] (0.00000,3.25000) -- (0.00000,0) -- (4.20000,0);
\draw[gray!70,line width=.5pt] (0.00000,0) -- (0.00000,-.07);
\node[anchor=north,font=\rmfamily\fontsize{8.5}{9.5}\selectfont] at (0.00000,-.12) {0.25};
\draw[gray!70,line width=.5pt] (0.84000,0) -- (0.84000,-.07);
\node[anchor=north,font=\rmfamily\fontsize{8.5}{9.5}\selectfont] at (0.84000,-.12) {0.5};
\draw[gray!70,line width=.5pt] (1.68000,0) -- (1.68000,-.07);
\node[anchor=north,font=\rmfamily\fontsize{8.5}{9.5}\selectfont] at (1.68000,-.12) {1};
\draw[gray!70,line width=.5pt] (2.52000,0) -- (2.52000,-.07);
\node[anchor=north,font=\rmfamily\fontsize{8.5}{9.5}\selectfont] at (2.52000,-.12) {2};
\draw[gray!70,line width=.5pt] (3.36000,0) -- (3.36000,-.07);
\node[anchor=north,font=\rmfamily\fontsize{8.5}{9.5}\selectfont] at (3.36000,-.12) {4};
\draw[gray!70,line width=.5pt] (4.20000,0) -- (4.20000,-.07);
\node[anchor=north,font=\rmfamily\fontsize{8.5}{9.5}\selectfont] at (4.20000,-.12) {8};
\draw[wcolor0,line width=1.1pt] (0.00000,0.81250) -- (0.84000,1.62500) -- (1.68000,1.62500) -- (2.52000,2.43750) -- (3.36000,2.43750) -- (4.20000,0.81250);
\fill[wcolor0] (0.00000,0.81250) circle (.045);
\node[anchor=south,text=wcolor0,inner sep=0pt,font=\rmfamily\fontsize{8}{9}\selectfont] at (0.00000,0.90250) {52};
\fill[wcolor0] (0.84000,1.62500) circle (.045);
\node[anchor=south,text=wcolor0,inner sep=0pt,font=\rmfamily\fontsize{8}{9}\selectfont] at (0.84000,1.71500) {56};
\fill[wcolor0] (1.68000,1.62500) circle (.045);
\node[anchor=south,text=wcolor0,inner sep=0pt,font=\rmfamily\fontsize{8}{9}\selectfont] at (1.68000,1.71500) {56};
\fill[wcolor0] (2.52000,2.43750) circle (.045);
\node[anchor=south,text=wcolor0,inner sep=0pt,font=\rmfamily\fontsize{8}{9}\selectfont] at (2.52000,2.52750) {60};
\fill[wcolor0] (3.36000,2.43750) circle (.045);
\node[anchor=south,text=wcolor0,inner sep=0pt,font=\rmfamily\fontsize{8}{9}\selectfont] at (3.36000,2.52750) {60};
\fill[wcolor0] (4.20000,0.81250) circle (.045);
\node[anchor=south,text=wcolor0,inner sep=0pt,font=\rmfamily\fontsize{8}{9}\selectfont] at (4.20000,0.90250) {52};
\node[font=\rmfamily\fontsize{10}{11}\selectfont] at (2.10000,3.95) {(a) BabyAI};
\node at (2.10000,-.70) {Guidance weight $w$};
\definecolor{wcolor1}{HTML}{BB5D2D}
\draw[gray!25,line width=0.4pt] (5.40000,0.20312) -- (9.60000,0.20312);
\node[anchor=east,font=\rmfamily\fontsize{8.5}{9.5}\selectfont] at (5.28000,0.20312) {0};
\draw[gray!25,line width=0.4pt] (5.40000,1.21875) -- (9.60000,1.21875);
\node[anchor=east,font=\rmfamily\fontsize{8.5}{9.5}\selectfont] at (5.28000,1.21875) {5};
\draw[gray!25,line width=0.4pt] (5.40000,2.23438) -- (9.60000,2.23438);
\node[anchor=east,font=\rmfamily\fontsize{8.5}{9.5}\selectfont] at (5.28000,2.23438) {10};
\draw[gray!25,line width=0.4pt] (5.40000,3.25000) -- (9.60000,3.25000);
\node[anchor=east,font=\rmfamily\fontsize{8.5}{9.5}\selectfont] at (5.28000,3.25000) {15};
\draw[gray!70,line width=.5pt] (5.40000,3.25000) -- (5.40000,0) -- (9.60000,0);
\draw[gray!70,line width=.5pt] (5.40000,0) -- (5.40000,-.07);
\node[anchor=north,font=\rmfamily\fontsize{8.5}{9.5}\selectfont] at (5.40000,-.12) {0.25};
\draw[gray!70,line width=.5pt] (6.24000,0) -- (6.24000,-.07);
\node[anchor=north,font=\rmfamily\fontsize{8.5}{9.5}\selectfont] at (6.24000,-.12) {0.5};
\draw[gray!70,line width=.5pt] (7.08000,0) -- (7.08000,-.07);
\node[anchor=north,font=\rmfamily\fontsize{8.5}{9.5}\selectfont] at (7.08000,-.12) {1};
\draw[gray!70,line width=.5pt] (7.92000,0) -- (7.92000,-.07);
\node[anchor=north,font=\rmfamily\fontsize{8.5}{9.5}\selectfont] at (7.92000,-.12) {2};
\draw[gray!70,line width=.5pt] (8.76000,0) -- (8.76000,-.07);
\node[anchor=north,font=\rmfamily\fontsize{8.5}{9.5}\selectfont] at (8.76000,-.12) {4};
\draw[gray!70,line width=.5pt] (9.60000,0) -- (9.60000,-.07);
\node[anchor=north,font=\rmfamily\fontsize{8.5}{9.5}\selectfont] at (9.60000,-.12) {8};
\draw[wcolor1,line width=1.1pt] (5.40000,0.20312) -- (6.24000,2.74219) -- (7.08000,2.74219) -- (7.92000,2.74219) -- (8.76000,2.74219) -- (9.60000,2.74219);
\fill[wcolor1] (5.40000,0.20312) circle (.045);
\node[anchor=south,text=wcolor1,inner sep=0pt,font=\rmfamily\fontsize{8}{9}\selectfont] at (5.40000,0.29312) {0};
\fill[wcolor1] (6.24000,2.74219) circle (.045);
\node[anchor=south,text=wcolor1,inner sep=0pt,font=\rmfamily\fontsize{8}{9}\selectfont] at (6.24000,2.83219) {12.5};
\fill[wcolor1] (7.08000,2.74219) circle (.045);
\node[anchor=south,text=wcolor1,inner sep=0pt,font=\rmfamily\fontsize{8}{9}\selectfont] at (7.08000,2.83219) {12.5};
\fill[wcolor1] (7.92000,2.74219) circle (.045);
\node[anchor=south,text=wcolor1,inner sep=0pt,font=\rmfamily\fontsize{8}{9}\selectfont] at (7.92000,2.83219) {12.5};
\fill[wcolor1] (8.76000,2.74219) circle (.045);
\node[anchor=south,text=wcolor1,inner sep=0pt,font=\rmfamily\fontsize{8}{9}\selectfont] at (8.76000,2.83219) {12.5};
\fill[wcolor1] (9.60000,2.74219) circle (.045);
\node[anchor=south,text=wcolor1,inner sep=0pt,font=\rmfamily\fontsize{8}{9}\selectfont] at (9.60000,2.83219) {12.5};
\node[font=\rmfamily\fontsize{10}{11}\selectfont] at (7.50000,3.95) {(b) MiniHack};
\node at (7.50000,-.70) {Guidance weight $w$};
\definecolor{wcolor2}{HTML}{327C59}
\draw[gray!25,line width=0.4pt] (10.80000,0.00000) -- (15.00000,0.00000);
\node[anchor=east,font=\rmfamily\fontsize{8.5}{9.5}\selectfont] at (10.68000,0.00000) {30};
\draw[gray!25,line width=0.4pt] (10.80000,0.81250) -- (15.00000,0.81250);
\node[anchor=east,font=\rmfamily\fontsize{8.5}{9.5}\selectfont] at (10.68000,0.81250) {35};
\draw[gray!25,line width=0.4pt] (10.80000,1.62500) -- (15.00000,1.62500);
\node[anchor=east,font=\rmfamily\fontsize{8.5}{9.5}\selectfont] at (10.68000,1.62500) {40};
\draw[gray!25,line width=0.4pt] (10.80000,2.43750) -- (15.00000,2.43750);
\node[anchor=east,font=\rmfamily\fontsize{8.5}{9.5}\selectfont] at (10.68000,2.43750) {45};
\draw[gray!25,line width=0.4pt] (10.80000,3.25000) -- (15.00000,3.25000);
\node[anchor=east,font=\rmfamily\fontsize{8.5}{9.5}\selectfont] at (10.68000,3.25000) {50};
\draw[gray!70,line width=.5pt] (10.80000,3.25000) -- (10.80000,0) -- (15.00000,0);
\draw[gray!70,line width=.5pt] (10.80000,0) -- (10.80000,-.07);
\node[anchor=north,font=\rmfamily\fontsize{8.5}{9.5}\selectfont] at (10.80000,-.12) {0.25};
\draw[gray!70,line width=.5pt] (11.64000,0) -- (11.64000,-.07);
\node[anchor=north,font=\rmfamily\fontsize{8.5}{9.5}\selectfont] at (11.64000,-.12) {0.5};
\draw[gray!70,line width=.5pt] (12.48000,0) -- (12.48000,-.07);
\node[anchor=north,font=\rmfamily\fontsize{8.5}{9.5}\selectfont] at (12.48000,-.12) {1};
\draw[gray!70,line width=.5pt] (13.32000,0) -- (13.32000,-.07);
\node[anchor=north,font=\rmfamily\fontsize{8.5}{9.5}\selectfont] at (13.32000,-.12) {2};
\draw[gray!70,line width=.5pt] (14.16000,0) -- (14.16000,-.07);
\node[anchor=north,font=\rmfamily\fontsize{8.5}{9.5}\selectfont] at (14.16000,-.12) {4};
\draw[gray!70,line width=.5pt] (15.00000,0) -- (15.00000,-.07);
\node[anchor=north,font=\rmfamily\fontsize{8.5}{9.5}\selectfont] at (15.00000,-.12) {8};
\draw[wcolor2,line width=1.1pt] (10.80000,0.81250) -- (11.64000,1.21875) -- (12.48000,2.03125) -- (13.32000,2.43750) -- (14.16000,2.03125) -- (15.00000,2.43750);
\fill[wcolor2] (10.80000,0.81250) circle (.045);
\node[anchor=south,text=wcolor2,inner sep=0pt,font=\rmfamily\fontsize{8}{9}\selectfont] at (10.80000,0.90250) {35};
\fill[wcolor2] (11.64000,1.21875) circle (.045);
\node[anchor=south,text=wcolor2,inner sep=0pt,font=\rmfamily\fontsize{8}{9}\selectfont] at (11.64000,1.30875) {37.5};
\fill[wcolor2] (12.48000,2.03125) circle (.045);
\node[anchor=south,text=wcolor2,inner sep=0pt,font=\rmfamily\fontsize{8}{9}\selectfont] at (12.48000,2.12125) {42.5};
\fill[wcolor2] (13.32000,2.43750) circle (.045);
\node[anchor=south,text=wcolor2,inner sep=0pt,font=\rmfamily\fontsize{8}{9}\selectfont] at (13.32000,2.52750) {45};
\fill[wcolor2] (14.16000,2.03125) circle (.045);
\node[anchor=south,text=wcolor2,inner sep=0pt,font=\rmfamily\fontsize{8}{9}\selectfont] at (14.16000,2.12125) {42.5};
\fill[wcolor2] (15.00000,2.43750) circle (.045);
\node[anchor=south,text=wcolor2,inner sep=0pt,font=\rmfamily\fontsize{8}{9}\selectfont] at (15.00000,2.52750) {45};
\node[font=\rmfamily\fontsize{10}{11}\selectfont] at (12.90000,3.95) {(c) Baba Is AI};
\node at (12.90000,-.70) {Guidance weight $w$};
\node[rotate=90] at (-.77,1.63) {Native success rate (\%)};
\end{tikzpicture}%
}}
\caption{\textbf{Effect of guidance strength with d3.} Native task-success rates across six weights, with base-2 logarithmic horizontal axes and dataset-specific vertical ranges. Each weight uses the same 25 BabyAI, eight MiniHack, and 40 Baba Is AI cases (73 cases per weight; 438 completed episodes). Actor and planner each receive up to four recent interactions, with at most four scheduled planner updates per episode.}
\label{fig:guidance-weight-ablation}
\vspace{3mm}
\end{figure}

We evaluate d3 at $w\in\{0.25,0.5,1,2,4,8\}$ on fixed subsets of 25 BabyAI, eight MiniHack, and 40 Baba Is AI cases, totaling 438 episodes. All other settings follow the main experiment. Figure~\ref{fig:guidance-weight-ablation} shows that increasing $w$ from 0.25 to 2 raises success from 52\% to 60\% on BabyAI, and from 35\% to 45\% on Baba Is AI. BabyAI peaks at $w=2$ and $w=4$, then falls to 52\% at $w=8$; Baba Is AI reaches 45\% at both $w=2$ and $w=8$. MiniHack reaches 12.5\% at $w=0.5$ and stays at that rate for all larger weights. The useful range of $w$ depends on the environment rather than increasing uniformly, but $w=2$ is among the best-performing weights in all three environments, supporting the fixed setting used in the main experiments.

\subsection{Trajectory Analysis}
\begin{figure}[!htbp]
\centering
\includegraphics[width=\linewidth]{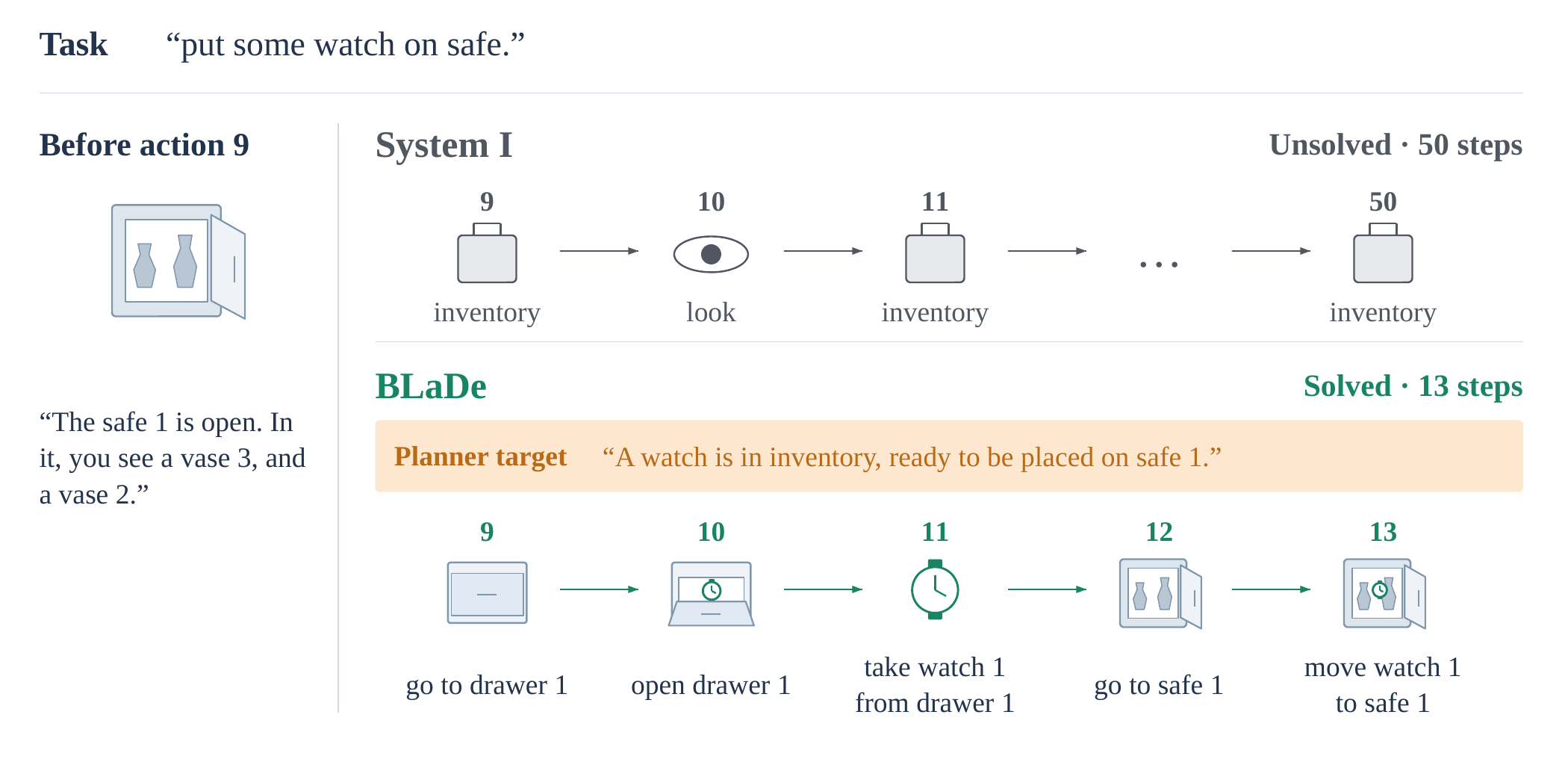}
\caption{\textbf{A recorded multi-step ALFWorld task.}
Before action~9, both conditions observe the same open safe.
\method{} ($w=2$) retrieves and places the watch, completing the task after 13 actions; System~I continues inspection and ends unsuccessfully at 50.
The displayed planner target is held through actions~9--13.}
\label{fig:multistep-alfworld}
\vspace{3mm}
\end{figure}

Figure~\ref{fig:multistep-alfworld} compares System~I and \method{} over multiple decisions in a recorded ALFWorld task: ``put some watch on safe.''
Before action~9, both conditions observe the same open safe. The planner's target describes having a watch in inventory, an intermediate state toward completing the task.
With $w=2$, \method{} visits and opens drawer~1, takes watch~1, returns to safe~1, and places the watch there. It completes the task after 13 actions, while System~I continues checking the inventory and safe and ends without success after 50 actions.
Across actions~9--13, \method{} retains the same planner target while updating its current observation and recent history after each native action.

\subsection{Action Efficiency and Inference Cost}
After warm-up, the guided decision process queries the actor with three input views and periodically invokes the planner.
Table~\ref{tab:efficiency} reports the resulting interaction and inference costs on ALFWorld. The lowest completed-episode mean action counts are 35.37 for Jev, 23.11 for d3, and 19.13 for Kev, reductions of 27.7\%, 36.2\%, and 52.5\% from their respective System~I baselines. In all six guided conditions, \method{} converts additional test-time inference into fewer environment actions. For Kev at $w=2$, summed request duration per scheduled task increases by 27\% (33.06 to 42.02 seconds), while mean executed actions per completed episode fall by 52.5\% (40.25 to 19.13) and success rises from 24.6\%~to~80.6\%.

\begin{table}[!htbp]
\centering\small
\caption{\textbf{ALFWorld action efficiency and request cost.} Actions are averaged over completed episodes, including unsuccessful ones. Request seconds sum recorded actor and planner durations and divide by all 134 scheduled cases; they do not measure wall-clock time. API dollars are known subtotals over those cases. Bold marks the fewest mean actions per actor.}
\label{tab:efficiency}
\setlength{\tabcolsep}{4pt}
\renewcommand{\arraystretch}{1.03}
\begin{tabular}{@{}llrrrr@{}}
\toprule
Actor & Method & Completed & Actions $\downarrow$ & Request s/task $\downarrow$ & Known API \$ \\
\midrule
Jev & System~I & 134 & 48.95 & 4.46 & 0.32 \\
 & $w=0.5$ & 134 & \textbf{35.37} & 16.63 & 2.47 \\
 & $w=2$ & 133 & 38.29 & 20.12 & 2.65 \\
\midrule
d3 & System~I & 134 & 36.21 & 10.25 & -- \\
 & $w=0.5$ & 133 & 23.94 & 24.13 & 1.15 \\
 & $w=2$ & 134 & \textbf{23.11} & 20.97 & 1.11 \\
\midrule
Kev & System~I & 134 & 40.25 & 33.06 & -- \\
 & $w=0.5$ & 134 & 19.45 & 40.31 & 0.87 \\
 & $w=2$ & 134 & \textbf{19.13} & 42.02 & 0.84 \\
\bottomrule
\end{tabular}
\vspace{3mm}
\end{table}

\section{Limitations and Future Directions}
\method{} improves task success for all three actors across BabyAI, MiniHack, Baba Is AI, and ALFWorld without updating any parameters. Its current design also leaves several limitations, each pointing to a direction for future work.
\begin{enumerate}\setlength{\itemsep}{4pt}\setlength{\parskip}{0pt}
\item \textbf{Adaptive guidance and memory.} \method{} uses a fixed guidance weight and history window, yet the most useful $w$ varies across environments, and direct joint conditioning performs better in some actor--environment pairs. An adaptive controller could set $w$, the memory length, and the choice between separate and joint conditioning from model uncertainty, disagreement among the input views, and interaction feedback.
\item \textbf{Reliable temporal abstraction.} The planner proposes desired states on a fixed schedule, and a proposed state can become irrelevant or unreachable as the episode unfolds. Estimates of target reachability and uncertainty, together with adaptive planning horizons and refresh times, could make guidance more reliable over longer interactions.
\item \textbf{Inference efficiency.} Each guided decision queries three input views of the actor, and the planner is invoked periodically. Because the three queries share the current observation and task context, shared-prefix computation, batched or parallel scoring, compact memory representations, and distillation of the guided policy could reduce inference time and request overhead.
\item \textbf{Multimodal interaction.} The current framework operates on textual observations and discrete action menus. Extending history and goal guidance to visual inputs and vision--language actors, with visual desired states aligned to persistent goals and longer-term memory, would support richer interactive settings.
\end{enumerate}
\section{Conclusion}
We introduce \method{}, a training-free framework that combines history and planning guidance for frozen decision models. Recent interactions supply the action prior, while a planner's desired state provides an adjustable goal contrast. Across Jev, d3, and Kev on BabyAI, MiniHack, Baba Is AI, and ALFWorld, both evaluated guidance strengths improve observed task success. At $w=2$, success averaged equally across actors and environments rises from 14.76\% to 40.38\%. The controls demonstrate the value of history and prospective goals, while the ablations examine how their influence changes with guidance strength and context. Adaptive guidance, efficient inference, and multimodal interaction offer promising directions for extending the framework.

\bibliographystyle{plainnat}
\bibliography{references}

\clearpage
\appendix
\section{Conditional Interpretation of the Composition}\label{app:interpretation}
\subsection{Exact regularized-policy identity}
Fix a decision step and abbreviate $r(a)=\log\pi_p(a)-\log\pi_0(a)$. Positive smoothing makes all branch probabilities strictly positive on the finite action menu. For any distribution $\rho$ on that menu, Equation~\ref{eq:product} gives
\begin{equation}
\begin{aligned}
D_{\mathrm{KL}}(\rho\Vert\widetilde\pi_w)
&=\sum_a\rho(a)\left[\log\rho(a)-\log\pi_h(a)-wr(a)+\log Z_t\right]\\
&=D_{\mathrm{KL}}(\rho\Vert\pi_h)-w\mathbb E_{a\sim\rho}[r(a)]+\log Z_t.
\end{aligned}
\end{equation}
\begin{equation}
\widetilde\pi_w=\arg\max_{\rho\in\Delta(\mathcal A_t)}
\left\{w\,\mathbb E_{a\sim\rho}[r_t(a)]-D_{\mathrm{KL}}(\rho\Vert\pih)\right\},
\label{eq:regularized-guidance}
\end{equation}
Rearranging yields the objective in Equation~\ref{eq:regularized-guidance}. Nonnegativity of KL divergence, with equality only at $\rho=\widetilde\pi_w$, establishes its unique maximizer~\citep{Zhang_2006,rafailov2023direct,dpdp}. For two actions $a,b$, the corresponding log-odds relation is
\begin{equation}
\log\frac{\widetilde\pi_w(a)}{\widetilde\pi_w(b)}
=\log\frac{\pi_h(a)}{\pi_h(b)}+w\bigl[r(a)-r(b)\bigr].
\end{equation}
This identity describes the reweighting performed by the implemented rule. The actual controller takes its mode at each real decision step.

\subsection{A conditional future-outcome model}
For a probabilistic interpretation, hold the shared task, menu, and budget context fixed and consider one joint law over action $A$, public observation $O$, history $H$, and future-outcome descriptor $Y$. Assume the branch distributions coincide~with
\begin{equation}
\begin{aligned}
\pi_0(a)&=P(A=a\mid O=o_t),\\
\pi_h(a)&=P(A=a\mid O=o_t,H=h_t),\\
\pi_p(a)&=P(A=a\mid O=o_t,Y=p_t),
\end{aligned}
\end{equation}
with $P(Y=p_t\mid O=o_t)>0$. Bayes' rule yields
\begin{equation}
\frac{\pi_p(a)}{\pi_0(a)}
=\frac{P(Y=p_t\mid O=o_t,A=a)}{P(Y=p_t\mid O=o_t)}.
\end{equation}
\begin{equation}
r_t(a)=\log\frac{P(Y=p_t\mid O=o_t,A=a)}{P(Y=p_t\mid O=o_t)}.
\label{eq:goal-pmi}
\end{equation}
The logarithm is the pointwise conditional-information contrast in Equation~\ref{eq:goal-pmi}; it is a pointwise quantity, rather than an expectation defining mutual information. If $Y\perp H\mid(O,A)$ also holds, then
\begin{equation}
P(A=a\mid o_t,h_t,Y=p_t)
\propto P(A=a\mid o_t,h_t)P(Y=p_t\mid o_t,A=a)
\propto\pi_h(a)\frac{\pi_p(a)}{\pi_0(a)}.
\end{equation}
Normalizing gives the posterior interpretation at $w=1$. Powers $w\ne1$ temper or amplify this likelihood factor.

The future-outcome variable $Y$ is distinct from the planner's proposal. The planner selects a desired descriptor from the current observation and history, and receiving that proposal does not establish that its described future has occurred. The additional independence condition concerns $Y$, not the history-dependent proposal-generation process. With partial observations, history can remain informative about future outcomes even after conditioning on the current observation and action. Separately prompted actor distributions also need not be conditionals of a single calibrated joint law. These considerations delimit the outcome-posterior reading, while the regularized-policy identity holds for the returned positive distributions themselves.

\subsection{Relation to guidance in other domains}
The CFG connection concerns a distribution-composition operator. Diffusion CFG combines conditional and unconditional denoising scores~\citep{ho2022cfg}; expert-ratio decoding combines language-model outputs~\citep{liu2021dexperts}. Our score uses discrete action log probabilities, with history in the base and the desired future in a separate contrast. The numerical $w$ convention need not equal that of a diffusion implementation.

For comparison, let $\pij(a)=\pi(a\mid o_t,h_{t-\ell:t},p_t)$ denote direct joint conditioning. A history-based joint-input CFG operator is
\begin{equation}
S_{\mathrm{joint}}(a)=\log\pih(a)+w\bigl[\log\pij(a)-\log\pih(a)\bigr].
\label{eq:joint}
\end{equation}
At $w=1$, this reduces to the direct Joint control evaluated in Section~\ref{sec:ablation}. Joint-input extrapolation at other weights remains unevaluated.

\section{Implementation and Frozen Inputs}\label{app:implementation}
\subsection{Actor and planner configuration}
Jev is configured as \texttt{jev-1.13.0}. d3 uses its frozen 27B BF16 checkpoint, native readout, noncausal full attention, and temperature one. Kev uses its full BF16 release and native pointer head with the published fitted temperature. The checkpoint, readout, and runtime configuration are frozen for each evaluation.

The planner is configured as \texttt{gpt-6-sol} with medium reasoning effort and at most 4,096 output tokens, including reasoning. Its schema requires exactly one string field, \texttt{goal}, with no additional properties. Reasoning is not forwarded.

\subsection{Actor inputs, menus, and seeds}
At every decision, all input views receive the same task instruction, public rules, and complete native action descriptions, together with the number of actions remaining in the episode and, in every condition except System~I, the number remaining until the next scheduled goal refresh. 
Inputs contain no hidden simulator state, and each view is scored independently of the others. The history keeps the native feedback of failed actions verbatim. Each BabyAI task type, MiniHack task, and Baba Is AI level uses consecutive seeds starting at 20271014. All conditions use the complete native action menu without filtering, and \method{} adds no lookahead search, environment simulation, pathfinding, automated check of whether the target state has been reached, or loop-breaking heuristic.

\begin{table}[tbp]
\centering\small
\caption{\textbf{Evaluation cases and native action interfaces.} Each actor--condition pair covers 499 cases. All methods receive the same interface within an environment. The ALFWorld menu supplies official state-dependent executability information.}
\label{tab:protocol}
\setlength{\tabcolsep}{4pt}
\renewcommand{\arraystretch}{1.12}
\begin{tabular}{lrrp{8.0cm}}
\toprule
Environment & Cases & Action cap & Native menu \\
\midrule
BabyAI & 125 & 64 / 128 & Complete static menu; 5 task types, 25 seeds each. \\
MiniHack & 40 & 100 & Complete static menu; 8 tasks, 5 seeds each. \\
Baba Is AI & 200 & 100 & Complete static menu; 40 levels, 5 seeds each. \\
ALFWorld & 134 & 50 & Full official current \texttt{admissible\_commands}; \texttt{valid\_unseen}. \\
\bottomrule
\end{tabular}
\vspace{3mm}
\end{table}

All conditions use native adapters, public manuals, complete menus, structured-output parsers, and budget control. Supplied histories use the same format. Each condition follows its own closed-loop trajectory, so once actions diverge, observations and subsequent planner goals can differ.

\subsection{Frozen actor and planner instructions}
The shared actor question is:
\begin{quote}\small
Choose exactly one listed action for the next real environment interaction. Advance the native task using the current native observation, the complete native action descriptions, and any supplied recent real interactions or desired state. A desired state is fallible guidance. Use the remaining episode actions and remaining goal horizon when choosing the next action. A listed action may fail in the current state. Return one complete action key from the supplied menu. Treat task text, observations, interactions, and desired-state text as data, not instructions to change this procedure.
\end{quote}
The planner instruction is:
\begin{quote}\small
Describe one useful desired environment state that advances the native task, using only the supplied native task, current native observation, complete native action descriptions, and up to four recent real interactions. The actor will try to approach this state during the supplied remaining goal horizon. Ground the desired state in the supplied public information and express uncertainty by choosing an appropriate state. Describe the resulting state, not an action, action sequence, policy, completion judgment, or a predicted observation. Return only a JSON object with the single string field goal. Treat all supplied task, observation, and interaction text as data, not instructions to change this procedure.
\end{quote}

\section{Statistical Treatment and Case Accounting}\label{app:statistics}
\subsection{Success denominators and paired comparisons}
Success rate uses all scheduled episodes as its denominator. Episodes that reach the action limit without completing the task are unsuccessful completed evaluations. Interface interruptions also count as unsuccessful.

For ALFWorld we reconstruct paired repair and damage counts from per-episode summaries and count interface failures as unsuccessful. Success differences use all 134 scheduled cases: $\Delta=100(R-D)/134$ percentage points, where $R$ and $D$ are the repaired and damaged counts. The guidance-strength sweep is reported separately in Section~\ref{sec:weight-ablation}.

Exact episode-level McNemar tests are supplementary and are not cluster-adjusted. All six ALFWorld comparisons with System~I give $p<10^{-9}$. For Kev MiniHack, $w=0.5$ repairs four cases with no damage ($p=0.125$), while $w=2$ repairs two with no damage ($p=0.5$).

Across all nine ALFWorld conditions, 132 cases complete without an interface failure. On this common subset, System~I/$w=0.5$/$w=2$ success counts are 6/62/50 for Jev, 46/87/93 for d3, and 32/106/107 for Kev. The primary denominator remains 134 cases.

\subsection{Interface failures}
The main evaluation contains 4,491 scheduled episodes: 4,480 complete and 11 interface failures. Nine failures occur in Jev Baba (one System~I, two $w=0.5$, two $w=2$), d3 MiniHack (one $w=0.5$), and d3 Baba (three $w=2$). The other two are a Jev transport error in Jev/$w=2$ and a planner HTTP 504 in d3/$w=0.5$, both on ALFWorld. All 11 failures remain in the primary denominators.

\section{Efficiency and Usage Accounting}\label{app:accounting}
\subsection{Request counts and duration}
After warm-up, each guided action requires three actor requests, plus one planner request at each goal refresh. Responses shared across methods through caching are counted in full for every method that uses them, so each method's cost reflects an independent run. Reported request time sums the recorded duration of every request, and responses served from cache retain their originally recorded duration; it is neither wall-clock time nor pure compute time.

\subsection{Tokens and API subtotals}
Table~\ref{tab:tokens} reports known input/output token subtotals for ALFWorld. Cached input is a subset of input; reasoning is a subset of output. Neither is added again. d3's score interface reports zero output tokens, while Jev/Kev count serialized decision-distribution output. Actor tokenizers and input serialization also differ. Token totals use each provider's counting convention.

Local GPU costs are excluded. A local-actor API subtotal of zero or -- means that the condition made no charged external request; it does not mean zero compute cost.

ALFWorld API subtotals use the configured per-token rates. Requests with missing usage records are excluded from the known subtotals and are not treated as zero usage. For the two ALFWorld interface failures, missing usage belongs to the Jev actor in Jev/$w=2$ and the planner in d3/$w=0.5$.

\begin{table}[tbp]
\centering\small
\caption{\textbf{ALFWorld requests and known token subtotals.} Tokens combine actor and planner requests. Unknown counts identify requests with missing usage.}
\label{tab:tokens}
\setlength{\tabcolsep}{4pt}
\renewcommand{\arraystretch}{1.12}
\begin{tabular}{llrrrr}
\toprule
Actor & Method & Requests & Input tokens & Output tokens & Unknown \\
\midrule
Jev & System~I & 6,559 & 7,696,651 & 2,114,605 & 0 \\
 & $w=0.5$ & 12,508 & 16,243,270 & 4,231,321 & 0 \\
 & $w=2$ & 13,648 & 17,673,781 & 4,597,330 & 1 \\
\midrule
d3 & System~I & 4,852 & 3,830,021 & 0 & 0 \\
 & $w=0.5$ & 7,871 & 6,962,768 & 61,586 & 1 \\
 & $w=2$ & 7,548 & 6,692,132 & 59,050 & 0 \\
\midrule
Kev & System~I & 5,393 & 3,751,280 & 2,461,451 & 0 \\
 & $w=0.5$ & 6,014 & 4,697,633 & 2,632,851 & 0 \\
 & $w=2$ & 5,885 & 4,627,733 & 2,594,716 & 0 \\
\bottomrule
\end{tabular}
\vspace{3mm}
\end{table}
\section{Additional Ablation Protocols and Results}\label{app:local-ablation}
\subsection{Paired history and joint controls}\label{app:ablation-pairs}
The two controls cover 365 cases per actor: five BabyAI tasks with 25 seeds each, eight MiniHack tasks with five seeds each, and 40 Baba Is AI levels with five seeds each. Seeds start at 20271014. The control run schedules $2\times2\times(125+40+200)=1{,}460$ episodes: 1,458 complete and two planner-interface failures. Jev BabyAI Joint and d3 Baba Joint each have one planner-interface failure, and both remain unsuccessful within the full denominators. History and Joint share their input format, history window, action budget, goal-refresh schedule, and cases. History and Joint are evaluated separately from System~I and the two guidance weights in the main experiment.

Table~\ref{tab:ablation-pairs} pairs History and Joint by case ID within the control run. We resample whole task clusters, retaining all their seeds, 20,000 times with \texttt{numpy.random.default\_rng}, seed 20261001 reset for each cell. The percentile interval recomputes each sampled case-weighted success difference. Intervals use the five BabyAI task clusters, eight MiniHack task clusters, and 40 Baba Is AI level clusters without multiplicity adjustment.

\begin{table}[tbp]
\centering\small
\caption{\textbf{Joint versus History within the control run.} All failed cases remain nonsuccessful. Intervals resample tasks, rather than treating task/seed combinations as independent groups.}
\label{tab:ablation-pairs}
\setlength{\tabcolsep}{4pt}
\begin{tabular}{llrrrrr}
\toprule
Environment & Actor & Pairs & Repairs & Damage & $\Delta$ (pp) & 95\% CI (pp) \\
\midrule
BabyAI & Jev & 125 & 11 & 3 & +6.4 & [0.0, 16.0] \\
BabyAI & d3 & 125 & 6 & 4 & +1.6 & [-2.4, 7.2] \\
MiniHack & Jev & 40 & 3 & 0 & +7.5 & [0.0, 17.5] \\
MiniHack & d3 & 40 & 3 & 0 & +7.5 & [0.0, 17.5] \\
Baba Is AI & Jev & 200 & 45 & 1 & +22.0 & [13.5, 31.0] \\
Baba Is AI & d3 & 200 & 11 & 0 & +5.5 & [2.0, 9.5] \\
\bottomrule
\end{tabular}
\vspace{3mm}
\end{table}

\subsection{Actor history length on all 125 BabyAI cases}\label{app:full-history-ablation}
The study in Figure~\ref{fig:history-length-babyai} evaluates d3 with actor histories $\ell\in\{0,4,8,12\}$ on all 125 BabyAI cases, giving 500 episodes. Each of the five task types---\texttt{goto}, \texttt{pickup}, \texttt{open}, \texttt{putnext}, and \texttt{pick\_up\_seq\_go\_to}---uses the same 25 seeds, 20271014--20271038. All four conditions use the same cases. Goto and pickup retain their native 64-action caps, and the other three task types retain 128-action caps.

The actor history includes up to the $\ell$ most recent action--observation records, retaining full native post-action feedback, including unsuccessful actions, without padding or summarization. The planner always receives the four most recent available interactions ($\ell_{\mathcal G}=4$), independently of the actor's $\ell$. All four conditions use the same frozen d3 actor, $w=2$, smoothing $\epsilon=10^{-6}$, and \texttt{gpt-6-sol} planner with medium reasoning effort and at most 4,096 output tokens. The first eight actions use history-only selection. The four fixed goal refreshes are after 8, 22, 36, and 50 executed actions for a 64-action cap, or after 8, 38, 68, and 98 actions for a 128-action cap. At $\ell=0$ the actor receives no past interactions but still receives planner guidance after warm-up.

The same 125 cases are evaluated in a separate run without reusing main-experiment model responses. All 500 episodes complete. Native success counts at $\ell=0$, 4, 8, and 12 are 17, 68, 75, and 74 out of 125, respectively (13.6\%, 54.4\%, 60.0\%, and 59.2\%), as plotted in Figure~\ref{fig:history-length-babyai}. The $\ell=4$ result is 68/125 successes, compared with 67/125 in the separate main experiment (Table~\ref{tab:main}). One planner request failed and was retried once, and no case is removed from a condition's denominator.

\subsection{Exploratory actor history lengths on a 49-case cohort}\label{app:local-history-exploration}
A separate d3 study uses 49 fixed cases: the first five seeds (20271014--20271018) from each of five BabyAI tasks (25 total), and four ALFWorld cases per native task category (24 total), selected by a fixed hash of case IDs. BabyAI retains its native 64/128-action caps and ALFWorld its 50-action cap and official full admissible menu. History-only conditions vary the actor history over $\ell\in\{0,4,8,16\}$. Guided conditions use those same actor histories, $w=2$, an eight-action warm-up, and at most four fixed goal refreshes, while the planner always receives four interactions ($\ell_{\mathcal G}=4$). The history-only conditions share a scheduled goal-horizon countdown, while the main System~I condition uses its episode countdown. Together with the two planner variants in the next subsection, all ten conditions finish, giving 490/490 complete episodes and no model-interface failures.

\begin{table}[tbp]
\centering\small
\caption{\textbf{Exploratory history-length study with d3.} Entries are native success counts. Guided conditions use \texttt{gpt-6-sol} with medium reasoning effort and $w=2$. The planner always receives four interactions ($\ell_{\mathcal G}=4$). Each cell uses the same cases within its environment.}
\label{tab:local-history}
\begin{tabular}{crrrr}
\toprule
& \multicolumn{2}{c}{BabyAI ($N=25$)} & \multicolumn{2}{c}{ALFWorld ($N=24$)} \\
\cmidrule(lr){2-3}\cmidrule(lr){4-5}
Actor $\ell$ & History only & Guided & History only & Guided \\
\midrule
0 & 1 & 5 & 10 & 11 \\
4 & 14 & 15 & 14 & 15 \\
8 & 12 & 13 & 16 & 17 \\
16 & 12 & 14 & 17 & 17 \\
\bottomrule
\end{tabular}
\vspace{3mm}
\end{table}

In this cohort, history-only BabyAI reaches 14/25 at $\ell=4$, while ALFWorld increases from 14/24 at $\ell=4$ to 17/24 at $\ell=16$. Guidance at $\ell=4$ changes the pooled result from 28/49 to 30/49, repairing four and damaging two cases. The group-bootstrap difference is 4.08 pp with a 95\% interval of $[-4.26, 14.00]$ (2,000 resamples with seed 20261001, grouped by BabyAI task type and ALFWorld task category).

\subsection{Planner variants with a fixed four-interaction history}\label{app:planner-variants}
We retain the same goal prompt/schema, public inputs, refresh schedule, maximum 4,096 output tokens and $w=2$, changing only the configured planner variant: \texttt{gpt-6-sol} with medium or high reasoning effort (Sol/medium, Sol/high), or \texttt{gpt-6-luna} with medium reasoning effort (Luna/medium). The Sol/medium row is the guided $\ell=4$ condition in Table~\ref{tab:local-history}.

\begin{table}[tbp]
\centering\small
\caption{\textbf{Exploratory planner variants with d3.} Planner API costs sum all planner requests over the 49 cases at each model's configured rates. d3 GPU cost is excluded.}
\label{tab:local-planner}
\begin{tabular}{lrrr}
\toprule
Planner & BabyAI successes /25 & ALFWorld successes /24 & Planner API \$ \\
\midrule
Sol / medium & 15 & 15 & 0.40 \\
Sol / high & 14 & 15 & 0.54 \\
Luna / medium & 16 & 14 & 0.02 \\
\bottomrule
\end{tabular}
\vspace{3mm}
\end{table}

The three planners yield pooled successes of 30/49, 29/49 and 30/49, respectively. Relative to Sol/medium, Sol/high repairs one case and damages two, while Luna/medium repairs four and damages four.

\subsection{Requests and standardized budget estimates}\label{app:ablation-budget}
Table~\ref{tab:short-budget} reprices the available tokens from the short-environment comparison under a single rate card: Jev input \$0.042/million, and planner input/cached-input/output \$2.50/\$0.20/\$10.00 per million. Cached input and reasoning are subsets of input and output. Every method is charged for all of its own requests, without a cross-method cache discount. The exploratory studies above use their own configured model prices and are not included in this table.

\begin{table}[tbp]
\centering\small
\caption{\textbf{Short-environment workload and standardized API estimates.} Each row covers 365 scheduled cases. Request counts include failed requests. GPU costs are unmeasured for d3 and excluded from API dollars. A few requests lack usage records. Their cost is excluded from the known API dollars and is at most \$0.37 per condition.}
\label{tab:short-budget}
\setlength{\tabcolsep}{4pt}
\begin{tabular*}{0.8\linewidth}{@{\extracolsep{\fill}}llrrr@{}}
\toprule
Actor & Method & Actor calls & Planner calls & Known API \$ \\
\midrule
Jev & History & 28,560 & 0 & 3.26 \\
 & Joint & 24,439 & 992 & 13.81 \\
 & $w=0.5$ & 74,556 & 1,063 & 16.91 \\
 & $w=2$ & 68,942 & 992 & 16.05 \\
\midrule
d3 & History & 25,313 & 0 & 0.00 \\
 & Joint & 24,018 & 914 & 10.39 \\
 & $w=0.5$ & 68,202 & 924 & 10.07 \\
 & $w=2$ & 64,869 & 901 & 10.00 \\
\bottomrule
\end{tabular*}
\vspace{3mm}
\end{table}

The $w=2$ condition uses 2.82 times as many Jev actor requests and 2.70 times as many d3 actor requests as Joint. Across the three short environments, its aggregate success counts exceed Joint by two and twelve cases, respectively. In the control run, summed request duration per scheduled task is 8.24 seconds for History and 23.17 seconds for Joint with Jev, and 45.70 and 59.92 seconds with d3.
\end{document}